\documentclass{article}

\usepackage{PRIMEarxiv}

\usepackage[utf8]{inputenc} % allow utf-8 input
\usepackage[T1]{fontenc}    % use 8-bit T1 fonts
\usepackage{hyperref}       % hyperlinks
\usepackage{url}            % simple URL typesetting
\usepackage{booktabs}       % professional-quality tables
\usepackage{amsfonts}       % blackboard math symbols
\usepackage{nicefrac}       % compact symbols for 1/2, etc.
\usepackage{microtype}      % microtypography
\usepackage{lipsum}
\usepackage{fancyhdr}       % header
\usepackage{graphicx}       % graphics
\graphicspath{{media/}}     % organize your images and other figures under media/ folder

\usepackage{amsmath}
\usepackage{graphicx}
\usepackage{amsmath}
\newtheorem{theorem}{Theorem}

\usepackage{algorithm}
\usepackage{algorithmic}
\usepackage{booktabs} %new
\usepackage{multirow}%new
\usepackage{amsmath}%new
\usepackage{amssymb}%new
\usepackage{pgfplots}%new
\usepackage{authblk}
\pgfplotsset{compat=1.18}
\title{Rhomboid Tiling for Geometric Graph Deep Learning
\thanks{\textit{\underline{Citation}}: 
\textbf{Accepted by ICML 2025}} 
}

\author{
Yipeng Zhang\textsuperscript{1,*},
Longlong Li\textsuperscript{1,2,3,*},
Kelin Xia\textsuperscript{1,\dag} \\
\textsuperscript{1}Division of Mathematical Sciences, School of Physical and Mathematical Sciences, Nanyang Technological University, Singapore 637371, Singapore \\
\textsuperscript{2}School of Mathematics, Shandong University, Jinan 250100, China \\
\textsuperscript{3}Data Science Institute, Shandong University, Jinan 250100, China \\
\texttt{yipeng001@e.ntu.edu.sg, longlee@mail.sdu.edu.cn, xiakelin@ntu.edu.sg}
}

\date{}

\begin{document}
\maketitle

\begin{abstract}
Graph Neural Networks (GNNs) have proven effective for learning from graph-structured data through their neighborhood-based message passing framework. Many hierarchical graph clustering pooling methods modify this framework by introducing clustering-based strategies, enabling the construction of more expressive and powerful models. However, all of these message passing framework heavily rely on the connectivity structure of graphs, limiting their ability to capture the rich geometric features inherent in geometric graphs. To address this, we propose Rhomboid Tiling (RT) clustering, a novel clustering method based on the rhomboid tiling structure, which performs clustering by leveraging the complex geometric information of the data and effectively extracts its higher-order geometric structures. Moreover, we design RTPool, a hierarchical graph clustering pooling model based on RT clustering for graph classification tasks. The proposed model demonstrates superior performance, outperforming 21 state-of-the-art competitors on all the 7 benchmark datasets.
\end{abstract}
%\keywords{Kolmogorov-Arnold Network, Fourier series, Graph Neural Network, Molecular Property Prediction}

\noindent\textsuperscript{*}Equal contribution. \\
\noindent\textsuperscript{\dag}Corresponding author.
%%\pacs[JEL Classification]{D8, H51}

%%\pacs[MSC Classification]{35A01, 65L10, 65L12, 65L20, 65L70}

\section{Introduction}
Graph Neural Networks (GNNs) have emerged as a powerful framework for learning from graph-structured data, which is pervasive in diverse domains such as social networks \cite{guo2020deep,min2021stgsn}, cheminformatics \cite{jiang2021could,kojima2020kgcn}, and computational biology \cite{li2021graph}. 
The foundation of GNNs lies in message passing, where nodes and edges exchange and aggregate message from their neighbors. 
This mechanism enables GNNs to capture both local and global relationships within the graph, extracting deep structural features for various graph-related tasks, such as node classification, link prediction, and graph classification. The versatility of GNNs enables them to address a range of real-world challenges, including identifying user roles or group memberships in social networks \cite{hamilton2017representation}, predicting molecular interactions or protein-protein interactions \cite{huang2020skipgnn,reau2023deeprank}, and predicting molecular properties \cite{wieder2020compact,shen2023molecular,cai2022fp}.

The message aggregation process is a fundamental component of GNNs, and the development of novel message aggregation mechanisms constitutes a critical direction for advancing GNN architectures. Traditional models such as GCN and GAT aggregate messages from a node’s neighbors and update the node’s representation based on the aggregated information. 
An alternative approach involves clustering nodes in the original graph into clusters, followed by aggregating messages within each cluster. A coarser graph is naturally generated through this process, with each cluster represented as a new node, enabling a hierarchical representation of the graph. This paradigm has led to the development of models referred to as hierarchical graph clustering pooling, such as DiffPool \cite{ying2018hierarchical} and MinCutPool \cite{bianchi2020spectral}. 
These models have shown considerable improvements in performance over traditional GNNs models. However, these clustering-based pooling methods primarily depend on the graph's connectivity structure and often utilize learnable matrices to determine node assignments to clusters, instead of leveraging predefined clusters derived from prior knowledge. In the case of geometric graphs, such as molecular graphs, the connectivity information alone may be insufficient to capture the rich geometric features inherent in the data. As a result, these methods face challenges in effectively incorporating geometric properties, limiting their ability to fully exploit the underlying structure of geometric graphs.

We propose a novel clustering method based on the rhomboid tiling structure for geometric graph deep learning. This method is inspired by the concept of Alpha shape, which is widely used in reconstructing 3D shapes from data, such as molecular surface reconstruction \cite{edelsbrunner1994three, liang1998analytical}. The core idea of Alpha shapes is to partition data using spheres, naturally enabling the construction of clustering methods. In fact, Alpha shapes have already been applied in clustering tasks in fields like cosmology \cite{gerke2012deep2}. Rhomboid tiling structures generalize Alpha shapes by capturing higher-order geometric information of point clouds while maintaining a natural hierarchical structure. Each layer of this structure corresponds to a high-order Delaunay complex (see Figure~\ref{fig:flowchart} \textbf{D}), where each vertex represents a substructure of the point cloud separated by a sphere from the remaining points. Consequently, each vertex can be regarded as a cluster of the point cloud. Moreover, the hierarchical nature of rhomboid tiling allows for further clustering of these clusters, enabling a hierarchical clustering method. This new clustering method is entirely driven by the geometric structure of the data and effectively uncovers complex geometric information that is difficult to extract using standard graph representations. Based on this method, we designed a graph clustering pooling model for graph classification tasks. 

Our main contributions are as follows:

1. We design a hierarchical clustering method, RT clustering, based on the rhomboid tiling structure to geometrically cluster data.

2. We provide a theoretical analysis of the optimal architectures for RT clustering and introduce a weighting mechanism to represent the importance of individual points within clusters.

3. We develop a graph clustering pooling model, RTPool, based on RT clustering and validate its performance on multiple graph classification tasks. Our model outperforms 21 state-of-the-art competitors on 7 benchmark datasets from chemistry and bioinformatics.

\section{Related Work}
\paragraph{Graph neural network and Graph Clustering Pooling} 
Graph Neural Networks (GNNs) have emerged as powerful tools for learning graph-structured data, with various architectures differing in their message aggregation mechanisms. The Graph Convolutional Network (GCN) aggregates node information using a neighborhood averaging approach weighted by normalized adjacency matrix entries, effectively capturing local smoothness \cite{kipf2017semisupervised}. The Graph Attention Network (GAT), on the other hand, employs attention mechanisms to learn the importance of neighboring nodes dynamically, allowing for more flexible aggregation based on node relationships. \cite{velivckovic2018graph} The Graph Isomorphism Network (GIN) utilizes sum-based aggregation, theoretically achieving maximal expressiveness among GNNs by distinguishing different graph structures \cite{xu2018powerful}. Another message aggregation mechanism involves clustering the nodes in a graph, performing message aggregation within each cluster, and updating the underlying graph into a coarsened graph where clusters act as nodes. Models based on this mechanism are referred to as graph clustering pooling models. DiffPool learns a differentiable cluster assignment matrix that maps nodes to clusters, jointly optimizing node representations and cluster assignments to generate a coarsened graph \cite{ying2018hierarchical}. MinCutPool, inspired by spectral clustering, performs pooling by minimizing the normalized cut of the graph while encouraging orthogonality in the cluster assignment matrix \cite{bianchi2020spectral}.
\paragraph{Voronoi Tessellation and Delaunay complex}
Voronoi tessellation and Delaunay complex are fundamental concepts in computational geometry \cite{fortune2017voronoi}. Voronoi tessellation partitions a space into regions around a set of points, such that each region consists of all points closer to its corresponding seed point than to any other. The Delaunay complex, on the other hand, is a dual structure of the Voronoi tessellation, comprising simplices formed by connecting points whose Voronoi cells share a common boundary. These two concepts have been widely applied across various domains due to their versatility and robustness in analyzing spatial relationships. For instance, in astronomy, Voronoi tessellation and Delaunay complexes are used to identify and characterize galaxy groups in large-scale cosmic surveys \cite{gerke2012deep2}. In material science, Voronoi-Delaunay analysis has been applied to study voids in systems of nonspherical particles, providing insights into structural properties of disordered systems \cite{luchnikov1999voronoi}. In protein structure analysis, it facilitates the computation of solvent-accessible surfaces and atomic packing densities \cite{richards1974interpretation}.

\section{Method}
\begin{figure*}[htbp]
    \centering
    \includegraphics[width=0.95\linewidth]{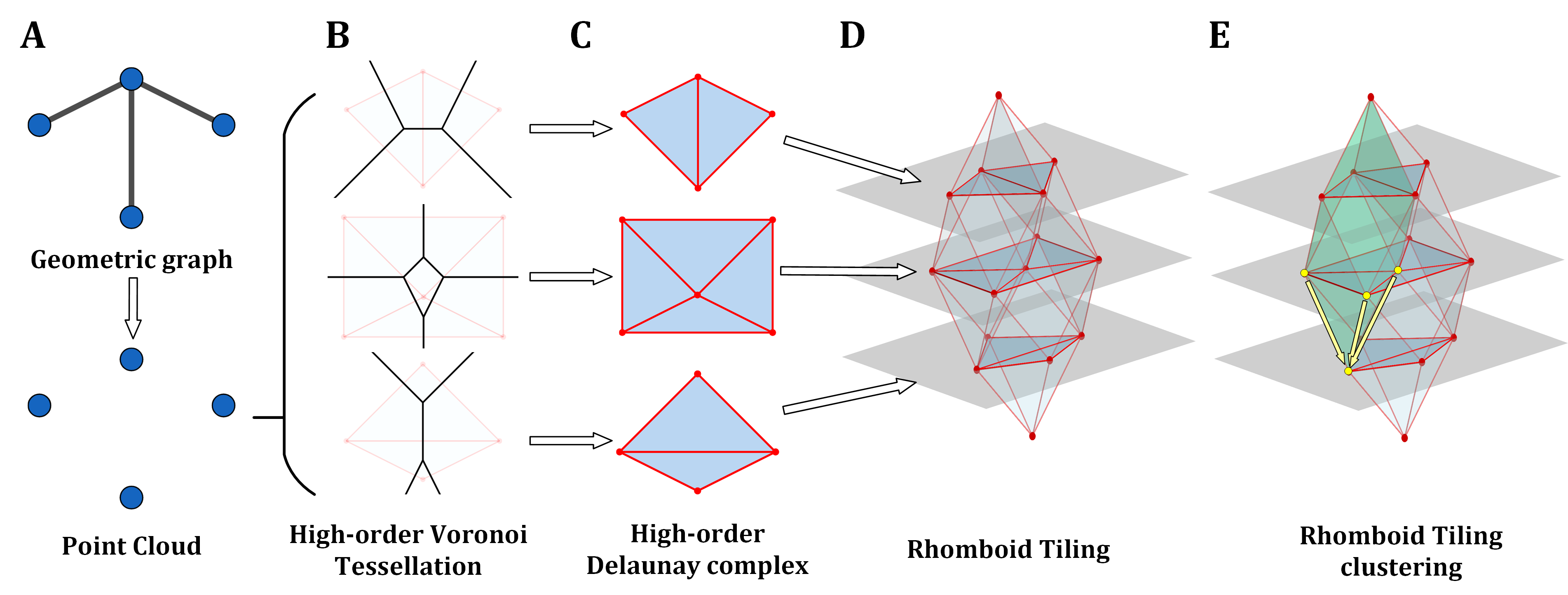}
    \caption{Flowchart of the Rhomboid Tiling clustering process. \textbf{A}: An example geometric graph and the corresponding point cloud \(X\), obtained by embedding the graph's vertices into \(\mathbb{R}^2\).
 \textbf{B}: 1-, 2-, and 3-order Voronoi tessellations constructed from \(X\). \textbf{C}: 1-, 2-, and 3-order Delaunay complexes obtained as the nerves of the corresponding Voronoi tessellations. \textbf{D}: The Rhomboid Tiling constructed based on \(X\). \textbf{E}: Illustration of 1-layer Rhomboid Tiling clustering, focusing on a single rhomboid.}

    \label{fig:flowchart}
\end{figure*}
\subsection{Rhomboid Tiling}
\paragraph{High-Order Voronoi Tessellation}  
High-order Voronoi tessellation generalizes the concept of the classical Voronoi tessellation, providing a method to partition space based on subsets of a given point set. For a point set \( X \subset \mathbb{R}^d \), consider each subset \( Q \subset X \). The Voronoi cell associated with \( Q \), denoted as \(\text{dom}(Q)\), is defined as:  
\[
\text{dom}(Q) = \{ p \in \mathbb{R}^d \mid \lVert p - x \rVert \leq \lVert p - y \rVert, \forall x \in Q, \forall y \in X \setminus Q \}.
\]

It is important to note that \(\text{dom}(Q)\) is non-empty only when the points in \( Q \)  can be separated from all other points in $X$ by a sphere. That is, there must exist a sphere \( S \) such that \( Q \) lies inside or on \( S \), while \( X \setminus Q \) lies outside \( S \). This implies that such subsets \( Q \) are, in some geometric sense, clustered together.  

All the subsets \( Q \subset X \) containing exactly \( k \) points define a partition of the entire space \( \mathbb{R}^d \), expressed as:  
\[
\mathbb{R}^d = \bigcup_{Q \subset X, |Q|=k} \text{dom}(Q).
\]

The collection of these partitions, denoted as  
\[
\text{Vor}_k(X) = \{\text{dom}(Q) \mid Q \subset X, |Q| = k, \text{dom}(Q)\neq \emptyset \},
\]  
is referred to as the \textit{order-\(k\) Voronoi tessellation}. This structure enables the exploration of higher-order relationships and geometric properties of point clusters within the space.

\paragraph{High-Order Delaunay Complex}  
Similar to how the traditional Delaunay complex is defined as the nerve of the classical Voronoi tessellation, the high-order Delaunay complex is defined as the nerve of the high-order Voronoi tessellation:  
\[
\text{Del}_k(X) := \text{Nrv}(\text{Vor}_k(X)).
\]
This means each vertex \(v_Q \in \text{Del}_k(X)\) corresponds to a cell \(\text{dom}(Q) \in \text{Vor}_k(X)\), which is associated with the subset \(Q \subset X\). A set of vertices \(v_{Q_1}, v_{Q_2}, \ldots, v_{Q_m}\) forms an \((m-1)\)-simplex in \(\text{Del}_k(X)\) if and only if the corresponding cells \(\text{dom}(Q_1), \text{dom}(Q_2), \ldots, \text{dom}(Q_m)\) have a non-empty intersection.  

If we interpret each subset \(Q_1, Q_2, \ldots, Q_m\) as a cluster, the simplex \((v_{Q_1}, v_{Q_2}, \ldots, v_{Q_m}) \in \text{Del}_k(X)\) signifies that these clusters are geometrically close. Thus, the high-order Delaunay complex \(\text{Del}_k(X)\) encodes the relationships between these clusters, allowing us to analyze high-order interactions within the given point set.

\paragraph{Rhomboid Tiling}  
To establish a natural relationship between Delaunay complexes of different orders, Edelsbrunner introduced the concept of Rhomboid Tiling \cite{edelsbrunner2021multi}. The core idea of this concept is to generalize the use of spheres for partitioning a given point set \( X \subset \mathbb{R}^d \) to relate Delaunay complexes of different orders. To achieve this, we first consider a way to describe the partition induced by a sphere \( S \). Let \(\text{In}_X(S)\), \(\text{On}_X(S)\), and \(\text{Out}_X(S)\) denote the subsets of \( X \) that are inside, on, and outside \( S \), respectively. Then each vertex \(v_Q \in \text{Del}_k(X)\) is mapped to \(\mathbb{R}^{d+1}\) using the following transformation:

\begin{equation}
\label{eq:trans}
    v_Q \mapsto y_Q := \left(\sum_{x \in Q} x, -k \right) \in \mathbb{R}^{d+1},
\end{equation}

It can be seen that we map the vertices of \(\text{Del}_k(X)\) to the hyperplane \(\{(x_1, x_2, \ldots, x_{d+1}) \in \mathbb{R}^{d+1} \mid x_{d+1} = -k\}\). The first \(d\) coordinates of \(y_Q\) are the sum of the coordinates of all points in the subset \(Q \subset X\). \(y_Q\) is the geometric realization of \(v_Q\) in \(\mathbb{R}^{d+1}\). Henceforth, unless otherwise specified or where ambiguity arises, we will not distinguish between \(v_Q\) and \(y_Q\).

Using the vertices from different \(\text{Del}_k(X)\) complexes and an arbitrary \((d-1)-\)dimensional sphere \( S \), a rhomboid constructed based on $S$ is defined as:  
\[
\rho_X(S) := \text{conv}\{y_Q \mid \text{In}_X(S) \subset Q \subset \text{In}_X(S) \cup \text{On}_X(S)\}.
\]

The collection of all such rhomboids forms a complex called the \textit{Rhomboid Tiling}:
\[
{\rm Rhomb}(X) := \{\rho_X(S) \mid S \text{ is a sphere in } \mathbb{R}^d\}.
\]

\text{Rhomb}(X) is a polyhedral complex. To ensure that this polyhedral complex exhibits good combinatorial properties, we assume that \(X\) is in general position in \(\mathbb{R}^d\). The concept of general position is commonly used in geometry and carries different meanings in various contexts. In this paper, we define \(X\) to be in general position in \(\mathbb{R}^d\) if no \(d+1\) points in \(X\) lie on the same \((d-1)\)-dimensional plane, and no \(d+2\) points lie on the same \((d-1)\)-dimensional sphere. All the data used in this study satisfy this condition.

Edelsbrunner proved that the intersection of \({\rm Rhomb}(X)\) with the hyperplane \(\{(x_1, x_2, \ldots, x_{d+1}) \in \mathbb{R}^{d+1} \mid x_{d+1} = -k\}\) corresponds precisely to the order-\(k\) Delaunay complex \(\text{Del}_k(X)\) when \(X\) is in general position in \(\mathbb{R}^d\) \cite{edelsbrunner2021multi}. Moreover, it can be proven that \({\rm Rhomb}(X) \cap \{(x_1, x_2, \ldots, x_{d+1}) \in \mathbb{R}^{d+1} \mid x_{d+1} \geq -k\}\) is homotopy equivalent to \(\text{Del}_k(X)\) \cite{corbet2023computing}. This indicates that if the portion of \({\rm Rhomb}(X)\) where the last coordinate is less than \(-k\) is discarded, the remaining part can be continuously deformed into \(\text{Del}_k(X)\).

To aid understanding, we provide a concrete example in the Appendix \ref{exp:rhomboidtiling} that illustrates how to construct a rhomboid tiling from a 2D point cloud.

\subsection{Rhomboid Tiling Clustering (RT clustering)}  
As previously mentioned, each vertex of \(\text{Del}_k(X)\) can be interpreted as a cluster of the point cloud \(X \subset \mathbb{R}^d\), where the cluster contains exactly \(k\) points. This cluster arises from partitioning the point cloud using spheres, ensuring that geometrically close points are grouped into the same cluster while avoiding the formation of excessive clusters in dense regions.  

This concept can be extended further. Instead of clustering the points in \(X\) to form the vertices of \(\text{Del}_{k_1}(X)\), we can define a second-level clustering on the vertex set of \(\text{Del}_{k_1}(X)\), such that the resulting clusters correspond bijectively to the vertices of \(\text{Del}_{k_2}(X) (k_2>k_1)\).  This second-level clustering is defined through the following relation:  
\[
v_Q \sim v_{Q'} \iff \exists \sigma \in \text{Rhomb}(X) \text{ such that } v_Q, v_{Q'} \in \sigma,
\]
where \(v_Q\) and \(v_{Q'}\) are vertices in \(\text{Del}_{k_1}(X)\) and \(\text{Del}_{k_2}(X)\,(k_2>k_1)\).  

Using this relation, we can define the cluster associated with a vertex \(v_{Q'} \) as:  
\[
C_{Q'} = \{v_Q \in \text{Del}_{k_1}(X) \mid v_Q \sim v_{Q'}\},
\]
where \(Q'\) is a subset of \(X\) such that \(v_{Q'} \in \text{Del}_{k_2}(X)\). In this way, the vertices in \(\text{Del}_{k_1}(X)\) are clustered into groups that correspond bijectively to the vertices in \(\text{Del}_{k_2}(X)\). Figure \ref{fig:flowchart} shows the flowchart of rhomboid tiling clustering.

The following theorem provides a necessary and sufficient condition for \(v_Q\) to belong to the cluster \(C_{Q'}\), offering a geometric explanation of the Rhomboid Tiling Clustering:

\begin{theorem}
\label{thm:basic}
    Vertice $v_Q\in \text{Del}_{k_1}(X)$ belongs to the cluster $C_{Q'}$ if and only if $\exists (d-1) \text{-dimensional sphere } S$ such that:
    \begin{itemize}
        \item $v_{Q'}\in \text{Del}_{k_2}(X)$ with $k_2>k_1,$
        \item $\text{In}_X(S)\subset Q\cap Q',$
        \item $Q\cup Q'\subset\text{In}_X(S)\cup \text{On}_X(S)$
    \end{itemize}
\end{theorem}

This theorem illustrates that the underlying idea of Rhomboid Tiling Clustering is similar to concepts like Alpha shapes, where spheres are used to partition a given point set \(X\). These partitions are then used to define geometric relationships among the points, enabling clustering of the points or further clustering of the already-formed clusters. This method ensures that points within the same cluster are close in the sense of geometric proximity, while the clustering structure also encodes higher-order geometric relationships within the point set \(X\).

Next, we discuss a more specific and practical scenario: the point set \(X\) lies in \(\mathbb{R}^3\) and is in general position. In this case, the following theorem provides guidance on how to choose appropriate values of \(k_1\) and \(k_2\) to cluster the points in \(\text{Del}_{k_1}(X)\) into the clusters corresponding to the points in \(\text{Del}_{k_2}(X)\).

\begin{theorem}
\label{thm:dim3}
    Suppose \(X\) is in general position in \(\mathbb{R}^3\). Considering the vertices in \(\text{Del}_{k_1}(X)\) and \(\text{Del}_{k_2}(X)\) with \(k_1 < k_2 \leq |X|\), we have:

    \textbf{Case 1:} \(k_2 - k_1> 4\), then for any \(v_Q \in \text{Del}_{k_1}(X)\), \(v_Q\) does not belong to any cluster corresponding to the vertices in \(\text{Del}_{k_2}(X).\)

    \textbf{Case 2:} \( 1 \leq k_2-k_1 \leq  2\), then for any \(v_Q \in \text{Del}_{k_1}(X)\), there exists \(v_{Q'} \in \text{Del}_{k_2}(X)\) such that \(v_Q \in C_{Q'}.\)

    \textbf{Case 3:} \(k_2 - k_1 = 1\), then for any \(Q \subset Q'\) with \(v_Q \in \text{Del}_{k_1}(X)\) and \(v_{Q'} \in \text{Del}_{k_2}(X)\), we have \(v_Q \in C_{Q'}.\)
\end{theorem}

\textbf{Note:} In the theorem above, there is one additional case that has not been explicitly discussed: \(3 \leq k_2 - k_1 \leq 4\). For this range of \(k_2-k_1\), we observe in our empirical data that there exist certain point clouds \(X\) and vertices \(v_Q \in \text{Del}_{k_1}(X)\) such that \(v_Q\) does not belong to any cluster corresponding to the vertices in \(\text{Del}_{k_2}(X).\)

This theorem indicates that the step size \(k_2 - k_1\) for clustering should be chosen within an appropriate range. From Theorem \ref{thm:dim3} Case 1, we know that \(k_2 - k_1\) cannot exceed 4, as otherwise, none of the points will be clustered into any cluster. Additionally, as noted above, \(k_2 - k_1\) should preferably not be 3 or 4, as in this case, some points may still fail to be clustered into any cluster. 

The ideal choice for \(k_2 - k_1\) is 1 or 2. According to Theorem \ref{thm:dim3} Case 2, in this range, all points are guaranteed to be clustered into at least one cluster. And the optimal choice might be \(k_2 - k_1 = 1\), as per Theorem \ref{thm:dim3} Case 3, where \(v_Q\) will be clustered into \(C_{Q'}\) as long as \(Q \subset Q'\). If we sequentially cluster the original information on every point in \(X\) into \(\text{Del}_{2}(X)\), then cluster \(\text{Del}_{2}(X)\) into \(\text{Del}_{3}(X)\), and so on, eventually reaching \(\text{Del}_{k}(X)\), the cluster \(C_{Q'}\) corresponding to a point in \(\text{Del}_{k}(X)\) will contain all the original information from points in \(Q' \subset X\).

\paragraph{Weight for RT clustering}
When performing RT clustering, introducing a weight for each point in a cluster \(C_{Q'}\) might be a beneficial choice. Although points \(v_{Q_1}\) and \(v_{Q_2}\) may belong to the same cluster \(C_{Q'}\), the geometric proximity or significance of the subsets \(Q_1\) and \(Q_2\) relative to \(Q'\) may differ. This motivates the introduction of a weight for each point \(v_Q\) in \(C_{Q'}\) to quantify such relationships. 

We propose using \(N(Q, Q') := \#\{\sigma \) is a depth-(d+1) rhomboid \(\mid v_Q, v_{Q'} \in \sigma\}\), which counts the number of depth-\((d+1)\) rhomboids simultaneously containing \(v_Q\) and \(v_{Q'}\), as a suitable weight to measure this relationship. Note that a depth-\(k\) rhomboid refers to a \(k\)-dimensional cell in \(\text{Rhomb}(X)\), where the corresponding sphere \(S\) satisfies \(|\text{On}_X(S)| = k\). A depth-\((d+1)\) rhomboid is the maximal rhomboid in \(\text{Rhomb}(X)\), assuming \(X\) is in general position in \(\mathbb{R}^d\).

The following theorem explains why \(N(Q, Q')\) is an appropriate choice for this weight:
\begin{theorem}
\label{thm:weight}
    Suppose \(X\) is in general position in \(\mathbb{R}^3\), and \(v_{Q_1}, v_{Q_2} \in \text{Del}_{k_1}(X),\, v_{Q'} \in \text{Del}_{k_2}(X)\). If \(v_{Q_1}, v_{Q_2} \in C_{Q'}\), then the following hold:
    \begin{enumerate}
        \item If \(3 \leq k_2 - k_1 \leq 4\), then \(Q_1 \subset Q'\) and \(N(Q_1, Q') \leq 5 - (k_2 - k_1)\).
        \item If \(Q_1 \not\subset Q'\), then \(N(Q_1, Q') \leq 3 - (k_2 - k_1)\).
        \item If \(Q_1 \cap Q' \subsetneq Q_2 \cap Q'\), then \(N(Q_1, Q') \leq N(Q_2, Q')\).
    \end{enumerate}
\end{theorem}

When \(|Q'| - |Q|\) is large or \(Q \not\subset Q'\), it can be inferred that \(Q\) and \(Q'\) are not geometrically closely connected. The first and second points of the theorem above indicate that, in such cases, \(N(Q, Q')\) does not exceed 2. 

Furthermore, when \(Q_1 \cap Q' \subsetneq Q_2 \cap Q'\), it is evident that \(Q_2\) has a closer geometric relationship with \(Q'\) than \(Q_1\) does. The third point of the theorem also confirms that, in this situation, \(N(Q_1, Q') \leq N(Q_2, Q')\). Thus, the theorem demonstrates that \(N(Q, Q')\) serves as a good metric for measuring the geometric relationship between \(Q\) and \(Q'\).

\subsection{RT clustering-based model} \label{RTmodel}
\paragraph{RT clustering-based pooling model (RTPool)}
It is a very natural idea to design a clustering pooling model based on RT clustering. We start by using the incident matrix \(I_k\) to represent which vertices of \(\text{Del}_k(X)\) are contained in which maximal rhomboids:

\[
(I_k)_{(i,j)} = 
\begin{cases} 
1, & \text{if } v_j \in \sigma_i, \\
0, & \text{otherwise}.
\end{cases}
\]

Here, \(\sigma_i\) denotes the \(i\)-th maximal rhomboid in \(\text{Rhomb}(X)\), and \(v_j\) represents the \(j\)-th vertex of \(\text{Del}_k(X)\). 

To cluster the vertices of \(\text{Del}_{k_1}(X)\) onto \(\text{Del}_{k_2}(X)\), we consider \(C_{k_1}^{k_2} := {(I_{k_2})}^T \cdot I_{k_1}\) as the clustering matrix. In fact, it is not difficult to prove that \((C_{k_1}^{k_2})_{(i,j)} = N(Q_i, Q_j')\), where \(v_{Q_i}\) is the \(i\)-th vertex of \(\text{Del}_{k_1}(X)\), and \(v_{Q_j}'\) is the \(j\)-th vertex of \(\text{Del}_{k_2}(X)\). In other words, the clustering matrix \(C_{k_1}^{k_2}\) defined in this way not only accounts for whether \(v_{Q_i}\) belongs to the cluster \(C_{Q_j'}\) but also incorporates the weight \(N(Q_i, Q_j')\) discussed in the previous section.

To construct a hierarchical pooling structure, we introduce a tunable hyperparameter \(\Delta k\) that specifies the step size in the order of the Delaunay complexes, i.e., we let \(k_2 - k_1 = \Delta k\). This means that at the \(l\)-th pooling layer (\(l = 0, 1, 2, \dots\)), node features are clustered from the order-\((l\Delta k + 1)\) Delaunay complex to the order-\(((l + 1)\Delta k + 1)\) Delaunay complex. Specifically, we use the matrix \(C_{l\Delta k+1}^{(l+1)\Delta k+1}\) for the clustering process. Each row of \(C_{l\Delta k+1}^{(l+1)\Delta k+1}\) is normalized by dividing each element by the sum of the elements in that row. The resulting row-normalized matrix is denoted as \(\hat{C}_l \in \mathbb{R}^{n_l \times n_{l+1}}\), where \(n_l\) and \(n_{l+1}\) are the numbers of vertices in \(\mathrm{Del}_{l\Delta k+1}(X)\) and \(\mathrm{Del}_{(l+1)\Delta k+1}(X)\), respectively. The normalized matrix \(\hat{C}_l\) is used as the clustering matrix for pooling from layer \(l\) to layer \(l+1\):

\begin{equation}
    Z^{(l+1)} = \hat{C_l} \cdot H^{(l)},
\end{equation}

where \(H^{(l)}\) is the node feature matrix at layer \(l\). Then we update the node features obtained after pooling using a given underlying graph \(G_{l+1}\) at layer \((l+1)\) and a GNNs model. This process produces the node features for the \((l+1)\)-th layer as follows:

\begin{equation}
    H^{(l+1)} = \text{GNNs}(Z^{(l+1)}, A_{l+1}),
    \label{eq:update}
\end{equation}

where \(A_{l+1}\) is the adjacency matrix of the underlying graph \(G_{l+1}\). And after pooling reaches the final layer \(L\), we compute the final embedding \(H_{\text{final}}\) using the following formula:

\[
    H_{\text{final}} = {H^{(L)}}^T \cdot (H^{(L)} W),
\]

where \(W\) is a learnable matrix of size \(f \times 1\), and \(f\) is the feature dimension. We use a 1-layer MLP to map the final embedding into a 2-dimensional vector, representing the classification scores for label 0 and label 1, respectively.

The underlying graph \(G_l\) used at layer \(l\) for updating the node feature can be defined arbitrarily, as long as its vertices correspond one-to-one with the vertices of \(\text{Del}_l(X)\). In this paper, we consider two methods for constructing \(G_l\):

1. The first method directly uses the 1-skeleton of \(\text{Del}_l(X)\) as \(G_l\). We refer to this type of underlying graphs as Delaunay graphs.

2. The second method assumes that the original point cloud \(X\) is associated with a given graph \(G_{ini}\). In our paper, \(G_{ini}\) is naturally defined as the corresponding chemical or molecular graph. The vertex set of \(G_l\) is given by the vertex set of \(\text{Del}_l(X)\), \(\{v_{Q_1}, v_{Q_2}, \dots\}\), where \(Q_1, Q_2, \dots\) correspond to subsets of \(X\) of size \(l\). An edge \((v_{Q_i}, v_{Q_j})\) exists in \(G_l\) if and only if there exist points \(p \in Q_i\) and \(q \in Q_j\) such that \((p, q)\) is an edge in the initial graph \(G_{ini}\). We refer to this type of underlying graphs as generated graphs.

\subsection{Time Complexity Analysis}

We theoretically analyze the time complexity of the proposed RTPool model. The following theorem characterizes the overall computational cost:

\begin{theorem}
Let \( K = \Delta k \cdot L + 1 \), where \( \Delta k \) is the step size and \( L \) is the number of pooling layers. Then the total time complexity of RTPool on a point cloud in $\mathbb{R}^d$ of size \( n \) is
\[
O\left(K^{\left\lceil \frac{d+3}{2} \right\rceil}n^{\left\lfloor \frac{d+1}{2} \right\rfloor}  + K^5 n^2 \right).
\]
The first term corresponds to the cost of constructing the rhomboid tiling up to order \( K \), and the second term accounts for the cumulative computation over all \( L \) pooling layers. Here, \( \lfloor \cdot \rfloor \) and \( \lceil \cdot \rceil \) denote the floor and ceiling functions, respectively.
\end{theorem}

In our experiments, the step size \( \Delta k \) and the number of pooling layers \( L \) are both set to at most 2, so the total order \( K = \Delta k \cdot L + 1 \) remains a small constant.  Moreover, all of the point clouds are in dimension $d=3$. Under this setting, the overall time complexity of RTPool simplifies to \( O(n^2) \), making it suitable for practical use.

Furthermore, empirical results in Appendix~\ref{sec:efficiency} support the theoretical analysis, demonstrating that RTPool achieves competitive efficiency compared to other state-of-the-art graph pooling methods.

\section{Experiments}
\subsection{Datasets}
We evaluate the performance of RTPool on graph classification tasks using seven real-world graph datasets from the commonly utilized TUDataset benchmark. Among these, three datasets represent chemical compounds, while the remaining four datasets are molecular compounds datasets.
\paragraph{Chemical Compound Datasets}  
The chemical compound datasets include COX-2 \cite{sutherland2003spline}, BZR \cite{sutherland2003spline}, and MUTAG \cite{debnath1991structure}. The \textbf{COX-2 dataset} comprises cyclooxygenase-2 (COX-2) inhibitors tested for their ability to inhibit the human recombinant COX-2 enzyme, with compounds classified as active or inactive. The \textbf{BZR dataset} includes ligands for the benzodiazepine receptor (BZR), with labels indicating activity or inactivity. The \textbf{MUTAG dataset} consists of chemical compounds categorized by their mutagenic effect on a specific bacterium, with labels distinguishing mutagenic from non-mutagenic compounds.  

\paragraph{Molecular Compound Datasets}  
The molecular compound datasets include PTC\_MM, PTC\_MR, PTC\_FM, and PTC\_FR \cite{chen2007chemdb}. These datasets classify chemical compounds based on their carcinogenicity in rodents: \textbf{PTC\_MM} (male mice), \textbf{PTC\_MR} (male rats), \textbf{PTC\_FM} (female mice), and \textbf{PTC\_FR} (female rats). Labels indicate whether a compound is carcinogenic or non-carcinogenic. 

For all datasets, compounds are represented as graphs where vertices correspond to atoms, and edges represent chemical bonds. Hydrogen atoms are removed during preprocessing.

\begin{table*}[htbp]
    \centering
        \caption{Performance of different models on benchmark datasets. The best performance for each dataset is highlighted in \textbf{bold}.}

    \begin{tabular}{lccccccc}
    \toprule
    \textbf{Model} & \textbf{BZR} & \textbf{COX2} & \textbf{MUTAG} & \textbf{PTC\_MR} & \textbf{PTC\_MM} & \textbf{PTC\_FM} & \textbf{PTC\_FR} \\ 
    No. graphs & 405 & 467 & 188 & 344 & 336 & 349 & 351  \\
No. avg nodes & 35.75  & 41.22 & 17.93 & 25.56 & 24.25 & 25.00 & 24.96 \\
    \midrule
    WL & 86.16±0.97 & 79.67±1.32 & 85.75±1.96 & 57.97±0.49 & 67.28±0.97 & 64.80±0.85 & 67.64±0.74 \\ 
    WL-OA & 87.43±0.81 & 81.08±0.89 & 86.10±1.95 & 62.70±1.40 & 66.60±1.16 & 66.28±1.83 & 67.82±5.03 \\ 
    HGK-WL & 81.42±0.60 & 78.16±0.00 & 75.51±1.34 & 59.90±4.30 & 67.22±5.98 & 64.72±1.66 & 67.90±1.81 \\ 
    HGK-SP & 81.99±0.30 & 78.16±0.00 & 80.90±0.48 & 57.26±1.41 & 57.52±9.98 & 52.41±1.79 & 66.91±1.46\\ 
    CSM & 84.54±0.65 & 79.78±1.04 & 87.29±1.25 & 58.24±2.44 & 63.30±1.70 & 63.80±1.00 & 65.51±9.82\\ 
    GCN & 79.34±2.43 & 76.53±1.82 & 80.42±2.07 & 62.26±4.80 & 67.80±4.00 & 62.39±0.85 & 69.80±4.40 \\ 
    DGCNN & 79.40±1.71 & 79.85±2.64 & 85.83±1.66 & 58.59±2.47 & 62.10±14.09 & 60.28±6.67 & 65.43±11.30 \\ 
    GIN & 85.60±2.00 & 80.30±5.17 & 89.39±5.60 & 64.60±7.00 & 67.18±7.35 & 64.19±2.43 & 66.97±6.17 \\ 
    SAGPool & 82.95±4.91 & 79.45±2.98 & 76.78±2.12 & 69.41±4.40 & 66.67±8.57 & 67.65±3.72 & 65.71±10.69 \\ 
    EigenGCN & 83.05±6.00 & 80.16±5.80 & 79.50±0.66 & N/A & N/A & N/A & N/A \\ 
    MinCutPool & 82.64±5.05 & 80.07±3.85 & 79.17±1.64 & 64.16±3.47 & N/A & N/A &  N/A \\ 
    Top-K & 79.40±1.20 & 80.30±4.21 & 67.61±3.36 & 64.70±6.80 & 67.51±5.96 & 65.88±4.26 & 66.28±3.71 \\ 
    DiffPool & 83.93±4.41 & 79.66±2.64 & 79.22±1.02 & 64.85±4.30 & 66.00±5.36 & 63.00±3.40 & 69.80±4.40 \\ 
    HaarPool & 83.95±5.68 & 82.61±2.69 & 90.00±3.60 & 66.68±3.22 & 69.69±5.10 & 65.59±5.00 & 69.40±5.21 \\ 
    PersLay & 82.16±3.18 & 80.90±1.00 & 89.80±0.90 & N/A & N/A & N/A & N/A \\ 
    MPR & N/A & N/A & 84.00±8.60 & 66.36±6.55 & 68.60±6.30 & 63.94±5.19 & 64.27±3.78\\ 
    FC-V & 85.61±0.59 & 81.01±0.88 & 87.31±0.66 & N/A & N/A & N/A & N/A \\ 
    SIN & N/A & N/A & N/A & 66.80±4.56 & 70.55±4.79 & 68.68±6.80 & 69.80±4.36 \\ 
    Wit-TopoPool & 87.80±2.44 & 87.24±3.15 & 93.16±4.11 & 70.57±4.43 & 79.12±4.45 & 71.71±4.86 & 75.00±3.51 \\
    HopPool &85.37±4.36 &85.11±3.74 &\textbf{94.74±4.76} &65.71±2.85 &73.59±5.27 &64.15±4.62 & 65.71±3.71 \\
    MvPool &78.05±3.38 &82.98±5.24  &  89.64±2.43&68.58±2.61 &70.65±4.83 &62.86±3.37 &65.72±2.14\\
    
    \midrule
    RTPool & \textbf{88.29±0.98} & \textbf{92.76±1.90} & \textbf{94.74±3.33} & \textbf{78.86±1.57} & \textbf{82.94±2.20} & \textbf{77.72±1.14} & \textbf{82.29±2.80} \\ \bottomrule
    \end{tabular}
    \label{table:main}
\end{table*}

\subsection{Baselines}  
Wit-TopoPool achieves state-of-the-art performance across the above datasets \cite{chen2023topological}, making it our primary baseline for comparison. To ensure a fair and consistent evaluation, we adopt the same seeds as Wit-TopoPool for a 90/10 random training/test split, guaranteeing identical training and test sets.

We evaluate the performance of RTPool by comparing it against 21 state-of-the-art methods across four categories: (I) Graph kernel-Based Methods, including (1) Weisfeiler–Lehman Kernel (WL) \cite{shervashidze2011weisfeiler}, (2) Weisfeiler–Lehman Optimal Assignment Kernel (WL-OA) \cite{kriege2016valid}, (3) Weisfeiler–Lehman Hash Graph Kernel (HGK-WL) \cite{morris2016faster}, (4) Shortest Path Hash Graph Kernel (HGK-SP) \cite{morris2016faster}, and (5) Subgraph Matching Kernel (CSM) \cite{kriege2012subgraph}; (II) Graph Neural Network-Based Methods, including (6) Graph Convolutional Network (GCN) \cite{kipf2017semisupervised}, (7) Deep Graph Convolutional Neural Network (DGCNN) \cite{zhang2018end}, and (8) Graph Isomorphism Network (GIN) \cite{xu2018powerful}; (III) Graph Pooling-Based Methods, including (9) Self-Attention Graph Pooling (SAGPool) \cite{lee2019self}, (10) GCNs with Eigen Pooling (EigenGCN) \cite{ma2019graph}, (11) Spectral Clustering Pooling (MinCutPool) \cite{bianchi2020spectral}, (12) TopKPooling with Graph U-Nets (Top-K) \cite{gao2019graph}, (13) Differentiable Pooling (DiffPool) \cite{ying2018hierarchical}, (14) Haar Graph Pooling (HaarPool) \cite{wang2020haar}, (15) Multi-hop Graph pooling (HopPool) \cite{zhang2024multi}, and (16) Multi-view Graph pooling (MvPool) \cite{ma2024graphadt}; and (IV) Topology-Based Methods, including (17) Neural Networks for Persistence Diagrams (PersLay) \cite{carriere2020perslay}, (18) Deep Graph Mapper (MPR) \cite{bodnar2021deep}, (19) Filtration Curves with a Random Forest (FC-V) \cite{o2021filtration}, (20) Message Passing Simplicial Networks (SIN) \cite{bodnar2021weisfeiler}, and (21) Witness complex-based topological pooling (Wit-TopoPool) \cite{chen2023topological}.
 
\subsection{Experiment Settings}
In our study, the experiments were conducted on a machine equipped with NVIDIA RTX A5000 GPUs with 32GB of memory. To enhance the model's performance across various datasets, we carefully selected appropriate hyperparameter settings, including learning rate, dropout ratio, and number of pooling layers, detailed in the Appendix Materials. The number of epochs was set to 500, and each dataset was evaluated five times, with the mean value used as the final metric and the standard deviation recorded. 

For the baseline results, the performance of HopPool and MvPool was obtained by running the original implementations with grid search to determine the best hyperparameters. For all other baseline methods, we directly cite the results from \cite{chen2023topological}, which also adopted grid search based on the settings specified in each original paper.

\subsection{Experiment Results}

The comparison of our model with 21 baseline methods across 7 benchmark datasets is summarized in Table \ref{table:main}. We also conducted an ablation study on the COX2, MUTAG, and PTC\_MR datasets to analyze the impact of replacing RTPooling with trivial mean pooling in the model, the effect of different GNNs models used for feature updates in the pooling layers, and the influence of the choice of the underlying graph during updates. The detailed results are presented in Table~\ref{table:pool}, \ref{table:gnn} and \ref{table:ug}. Furthermore, we provide additional experimental analyses in Appendix, including (i) hyperparameter sensitivity studies, (ii) graph regression task performance, (iii) social network dataset evaluation, and (iv) computational efficiency analysis. In all these tables, \text{N/A} indicates that the result is not available. Values are reported as mean accuracy ± standard deviation.

\paragraph{Molecular and chemical graph}
Table \ref{table:main} presents the performance comparison among 21 baseline models on the BZR, COX-2, MUTAG, and the four PTC datasets (PTC MR, PTC MM, PTC FM, and PTC FR) for graph classification tasks.  The performance results for all other models are directly from \cite{chen2023topological}. To ensure a fair comparison of results, we follow the same train-test split strategy used in their work, utilize the corresponding molecular or chemical compound graphs as the initial graphs, and adopt the same node features: simple 1-hot vectors encoding the atom types represented by the nodes. Our RTPool consistently outperforms all baseline models across these 7 datasets, with the runner-up being Wit-TopoPool in all cases. Notably, RTPool achieves an average relative improvement of 5.94\% over the runner-up, demonstrating its effectiveness. This significant performance gain highlights the limitations of existing methods. All baseline models are constrained by their reliance on graph structures and do not fully exploit the geometric information inherent in molecular or compound graphs. Our RTPool model employs a rhomboid tiling structure that is sensitive to geometric information. It performs well in detecting subtle geometric variations in densely packed regions of the point cloud, enabling it to fully leverage this geometric data. As a result, our model can make full use of this geometric information and consistently outperforms all the baseline approaches.

\paragraph{Ablation Study}
To assess the impact of each component in our RTPool model, we conducted comprehensive ablation experiments across multiple datasets, including COX2, MUTAG, and PTC\_MR, to gain deeper insights into their individual contributions. 
We designed three experiments for the ablation study: In the first experiment, we replaced the entire pooling component of our RTPool model with the commonly used mean pooling and max pooling. This experiment was conducted to validate the effectiveness of RTPool as the pooling mechanism in graph deep learning. In the second experiment, we used GCN, GAT, and GIN as the models for updating node features after pooling, as described in Equation \ref{eq:update}. This allowed us to explore the impact of employing different GNNs architectures for feature updates. Lastly, in the third experiment, we considered two types of graphs: Delaunay graphs and generated graphs. These graphs were used as the underlying structures for updating node features at each new layer after RT clustering pooling (see equation \ref{eq:update}). This experiment aimed to investigate how the choice of the underlying graph affects the performance of our model.

\begin{table}[htbp]
    \centering
    \caption{Performance comparison of replacing RTPooling with trivial mean pooling. The best performance for each dataset is highlighted in \textbf{bold}}
    \begin{tabular}{lcc}
    \toprule
    \textbf{Dataset} & \textbf{Pooling Method} & \textbf{Accuracy} \\ 
    \midrule
    \multirow{3}{*}{COX2} & mean pool & 78.72±0.85 \\
    & max pool & 78.31±0.81 \\
                          & RTPool & \textbf{92.76±1.90} \\
    \midrule
    \multirow{3}{*}{MUTAG} & mean pool & 84.21±1.37 \\
    & max pool & 75.53±2.26 \\
                           & RTPool & \textbf{94.74±3.33} \\
    \midrule
    \multirow{3}{*}{PTC\_MR} & mean pool & 65.71±0.00 \\
    & max pool & 67.43±1.40 \\
                             & RTPool & \textbf{78.86±1.57} \\ 
    \bottomrule
    \end{tabular}
    \label{table:pool}
\end{table}

The results in Table \ref{table:pool} highlight significant advantage of RTPool over trivial mean and max pooling across all three datasets. Our model achieved relative accuracy gains of 10\% to 20\%, underscoring the value of an advanced pooling mechanism. This substantial performance gap demonstrates RTPool's effectiveness in capturing both geometric and topological information, establishing it as a robust graph pooling model that consistently drives superior performance across tasks.

\begin{table}[htbp]
    \centering
    \caption{Performance of different GNNs models used for node feature updates. The best performance for each dataset is highlighted in \textbf{bold}.}
    \begin{tabular}{lcc}
    \toprule
    \textbf{Dataset} & \textbf{GNNs Model} & \textbf{Accuracy} \\ 
    \midrule
    \multirow{3}{*}{COX2} & GCN & 88.93±0.84 \\
                          & GAT & 87.23±1.15 \\
                          & GIN & \textbf{92.76±1.90} \\
    \midrule
    \multirow{3}{*}{MUTAG} & GCN & 89.47±2.33 \\
                          & GAT & 88.73±1.48 \\
                          & GIN & \textbf{94.74±3.33} \\
    \midrule
    \multirow{3}{*}{PTC\_MR} & GCN & 74.28±0.86 \\
                          & GAT & 71.43±1.40 \\
                          & GIN & \textbf{78.86±1.57} \\
    \bottomrule
    \end{tabular}
    \label{table:gnn}
\end{table}

Table \ref{table:gnn} demonstrates that while the choice of GNN model for feature updates impacts performance, the differences are relatively modest. For example, on the COX2 dataset, the accuracy gap between the top-performing model (GIN) and others (GCN, GAT) is around 4\%, increasing to about 5\% on the MUTAG and PTC\_MR datasets. Notably, GIN consistently achieves the highest accuracy across all datasets, likely due to its expressiveness and alignment with the Weisfeiler-Lehman graph isomorphism test, which enhances its effectiveness in graph classification tasks. These results suggest that GIN is the optimal choice for node feature updates in the RTPool model.

\begin{table}[htbp]
    \centering
    \caption{Performance comparison based on the choice of the underlying graph for feature updates. The best performance for each dataset is highlighted in \textbf{bold}.}
    \begin{tabular}{lcc}
    \toprule
    \textbf{Dataset} & \textbf{Underlying Graph} & \textbf{Accuracy} \\ 
    \midrule
    \multirow{2}{*}{COX2} & Delaunay & 88.94±0.85 \\
                          & generated & \textbf{92.76±1.90} \\
    \midrule
    \multirow{2}{*}{MUTAG} & Delaunay & 86.32±2.58 \\
                           & generated & \textbf{94.74±3.33} \\
    \midrule
    \multirow{2}{*}{PTC\_MR} & Delaunay & 71.43±0.00 \\
                             & generated & \textbf{78.86±1.57} \\ 
    \bottomrule
    \end{tabular}
    \label{table:ug}
\end{table}

Table \ref{table:ug} summarizes the third experiment's results, showing the impact of different underlying graph constructions on model performance. As discussed in Section \ref{RTmodel}, we explored two approaches for constructing underlying graphs after pooling: Delaunay graphs and generated graphs. The table highlights a performance gap of up to 8\% on the MUTAG dataset, underscoring the importance of graph choice. This gap arises because RT clustering already captures geometric structure, so Delaunay graphs provide little additional information, as many Delaunay-connected nodes are already clustered together. Moreover, RT clustering ignores the edge semantics of the input graph, which are critical in domains like molecular modeling where chemical bonds encode key functional information. By reconstructing post-pooling graphs using the original graph’s connectivity, we preserve high-order graph structural details and better integrate both geometric and graph connectivity information. This explains why using generated graphs consistently leads to better performance.

\paragraph{Hyperparameter Sensitivity Analysis}
Theorem~\ref{thm:dim3} suggests that the difference $\Delta k = k_2 - k_1$ between consecutive clustering levels should be set to 1 or 2, as larger values may result in some nodes being left unclustered, leading to information loss during pooling.

To validate this theoretical insight, we conduct a sensitivity analysis on $\Delta k$. The results, shown in Table~\ref{table:delta_k}.

\begin{table}[htbp]
    \centering
    \caption{Sensitivity of the hyperparameter $\Delta k = k_2-k_1$. The best performance for each dataset is highlighted in \textbf{bold}.}
    \begin{tabular}{lcc}
    \toprule
    \textbf{Dataset} & $\Delta k$ & \textbf{Accuracy} \\ 
    \midrule
    \multirow{3}{*}{COX2} & 1 & 89.36±2.33 \\
                          & 2 & \textbf{92.76±1.90} \\
                          & 3 & 86.38±1.90 \\
    \midrule
    \multirow{3}{*}{MUTAG} & 1 & \textbf{94.74±3.33} \\
                           & 2 & 89.64±2.36 \\
                           & 3 & 88.42±2.10 \\
    \midrule
    \multirow{3}{*}{PTC\_MR} & 1 & 76.57±1.14 \\
                             & 2 & \textbf{78.86±1.57} \\
                             & 3 & 69.71±1.56 \\ 
    \bottomrule
    \end{tabular}
    \label{table:delta_k}
\end{table}

We observe that RTPool achieves robust and competitive results when $\Delta k = 1$ or $2$, aligning well with the geometric guarantees of Theorem~\ref{thm:dim3}. In contrast, $\Delta k = 3$ leads to consistent performance drops across all datasets, likely due to some nodes failing to be grouped into valid clusters. We thus recommend setting $\Delta k = 1$ as the default choice, with $\Delta k = 2$ being a viable alternative that can yield even better results on some datasets. Additional sensitivity experiments on other hyperparameters are provided in Appendix~\ref{HSA}.

\section{Conclusion}
This paper introduces Rhomboid Tiling (RT) clustering, a hierarchical method designed to capture intricate high-order geometric information from geometric graphs. Building on this foundation, we developed RTPool, a pooling model that achieves exceptional performance across various tasks. In the future, we plan to extend RT clustering to topological deep learning models.

\bibliographystyle{unsrt}  
\bibliography{references}

\newpage
\appendix
\onecolumn
\section{Proofs of Theorems}
\begin{theorem}
\label{apxthm:basic}
    A vertex \(v_Q \in \text{Del}_{k_1}(X)\) belongs to the cluster \(C_{Q'}\) if and only if there exists a \((d-1)\)-dimensional sphere \(S\) such that:
    \begin{itemize}
        \item \(v_{Q'} \in \text{Del}_{k_2}(X)\) with \(k_2 > k_1,\)
        \item \(\text{In}_X(S) \subset Q \cap Q',\)
        \item \(Q \cup Q' \subset \text{In}_X(S) \cup \text{On}_X(S).\)
    \end{itemize}
\end{theorem}

\textbf{Proof.} From the definition of the rhomboid associated with \(S\), we have \(\text{In}_X(S) \subset Q \subset \text{In}_X(S) \cup \text{On}_X(S)\) and \(\text{In}_X(S) \subset Q' \subset \text{In}_X(S) \cup \text{On}_X(S)\). Using these properties, it follows directly that the stated conditions are sufficient and necessary for \(v_Q \in C_{Q'}.\)

\begin{theorem}
\label{apxthm:dim3}
    Suppose \(X\) is in general position in \(\mathbb{R}^3\). Considering the vertices in \(\text{Del}_{k_1}(X)\) and \(\text{Del}_{k_2}(X)\) with \(k_1 < k_2 \leq |X|\), we have:

    \textbf{Case 1:} \(k_2 - k_1> 4\), then for any \(v_Q \in \text{Del}_{k_1}(X)\), \(v_Q\) does not belong to any cluster corresponding to the vertices in \(\text{Del}_{k_2}(X).\)

    \textbf{Case 2:} \( 1 \leq k_2-k_1 \leq  2\), then for any \(v_Q \in \text{Del}_{k_1}(X)\), there exists \(v_{Q'} \in \text{Del}_{k_2}(X)\) such that \(v_Q \in C_{Q'}.\)

    \textbf{Case 3:} \(k_2 - k_1 = 1\), then for any \(Q \subset Q'\) with \(v_Q \in \text{Del}_{k_1}(X)\) and \(v_{Q'} \in \text{Del}_{k_2}(X)\), we have \(v_Q \in C_{Q'}.\)
\end{theorem}

\textbf{Proof:}
\begin{itemize}
    \item \textbf{Case 1:} \(k_2-k_1 > 4\). Actually we can prove a more general case: Suppose \(X\) is in general position in \(\mathbb{R}^d\), then for any vertex $v_Q\in \text{Del}_{k_1}(X)$ and any vertex $v_{Q'}\in\text{Del}_{k_2}(X)$ with $k_2\geq k_1+d+1,\, v_Q\notin C_{Q'}$:
    
    From Theorem \ref{apxthm:basic}, we know that \(Q \cup Q' - Q \cap Q' \subset Q \cup Q' - \text{In}_X(S) \subset \text{On}_X(S)\). When \(k_2 > k_1 + d + 1\), we have \(|\text{On}_X(S)| \geq |Q \cup Q' - Q \cap Q'| \geq k_2 - k_1 > d + 1\). This contradicts the general position assumption, which states that for any \((d-1)\)-dimensional sphere \(S \subset \mathbb{R}^d\), there can be at most \(d+1\) vertices of \(X\) exactly on \(S\). Therefore, no such cluster exists in this case.

    \item \textbf{Case 2:} \( 1 \leq k_2-k_1 \leq  2\).  
    Since \(v_Q \in \text{Del}_{k_1}(X)\), there exists a sphere \(S_0 \subset \mathbb{R}^d\) such that \(\text{In}_X(S_0)\subset Q \subset \text{In}_X(S_0) \cup \text{On}_X(S_0)\). Our goal is to find additional points \(v_1, \ldots, v_m\) (\(1 \leq m \leq 2\)) and construct a sphere \(S_m\) such that:
    \begin{itemize}
        \item \(S_0 \subset D_m\), here we use $D_m$ to denote the ball bounded by the sphere $S_m$.
        \item \(v_1, \ldots, v_m\) lie exactly on \(S_m,\)
        \item All other points in \(X - Q - \{v_1, \ldots, v_m\}\) do not lie inside \(S_m\).
    \end{itemize}
    If such points \(v_1, \ldots, v_m\) and sphere \(S_m\) can be found, we define \(Q' = Q \cup \{v_1, v_2, \ldots, v_m\}\). Then, it follows that \(\text{In}_X(S_m) \subset Q \cap Q'\) and \(Q \cup Q' \subset \text{In}_X(S_m) \cup \text{On}_X(S_m)\). By Theorem \ref{apxthm:basic}, we can conclude that \(v_Q \in C_{Q'}\).

    - \textbf{Subcase 2.1: } \(m = 1\).
      Let \(O\) and \(R\) denote the center and radius of \(S_0\), respectively. For each point \(v \in X - Q\), calculate its Euclidean distance to \(O\). Select the point \(v_1\) that is closest to \(O\), and define \(S_1\) as the sphere centered at \(O\) with radius \(|l(O, v_1)|\), where \(l(O_v, v_1)\) denotes the line segment between the points \(O_v\) and \(v_1\); this notation will be used to represent the line segment between any two points throughout the proof. It is straightforward to verify that \(S_1\) satisfies:
      \begin{itemize}
        \item \(S_0 \subset D_1,\)
        \item \(v_1\) lies exactly on \(S_1\),
        \item All other points in \(X - Q - \{v_1\}\) do not lie inside \(S_1\).
      \end{itemize}

    - \textbf{Subcase 2.2:} \(m = 2\).  
      First, construct \(v_1\) and \(S_1\) as in the previous case. Next, for each \(v \in X - Q - \{v_1\}\), consider the perpendicular bisector of \(v_1\) and \(v\), and find its intersection \(O_v\) with the line passing through \(O\) and \(v_1\). Let \(|l(O_v, v_1)|\) denote the distance between \(O_v\) and \(v_1\). Select the point \(v_2\) such that \(|l(O_v, v_1)|\) is minimized. Define \(S_2\) as the sphere centered at \(O_{v_2}\) with radius \(|l(O_{v_2}, v_1)|\). Then:
      \begin{itemize}
        \item \(v_1\) and \(v_2\) lie exactly on \(S_2\).\\
              It is obvious by the construction of \(S_2\).
        \item \(S_1\) is contained within \(D_2\).\\
              To demonstrate this, observe that \(O_{v_2}\), \(O\), and \(v_1\) are collinear, and the radii of the spheres \(S_2\) and \(S_1\) satisfy the inequality \(|l(O_{v_2}, v_1)| \geq |l(O, v_1)|\). This ensures that, apart from the common point \(v_1\), the sphere \(S_1\) is entirely enclosed within \(S_2\).
        
        \item All other points in \(X - Q - \{v_1, v_2\}\) do not lie inside \(S_2\).\\
              To prove this, note that by construction, for any \(v \in X - Q - \{v_1, v_2\}\), \(O_v\) lies on the perpendicular bisector of \(v_1\) and \(v\). Hence, the angle \(\alpha(O_v, v_1, v)\) satisfies \(\alpha(O_v, v_1, v) = \alpha(O_v, v, v_1)\) (Here \(\alpha(O_v, v_1, v)\) denotes the angle between the line segments \(v_1O_v\) and \(v_1v\); this notation will be used to represent the angle between any two line segments throughout the proof). Since \(|l(O_v, v_1)| \geq |l(O_{v_2}, v_1)|\), we have \(\alpha(O_{v_2}, v, v_1) \leq \alpha(O_v, v, v_1)=\alpha(O_v, v_1, v)=\alpha(O_{v_2}, v_1, v)\). Since the greater side of a triangle is opposite the greater angle, \(|l(O_{v_2}, v)| \geq |l(O_{v_2}, v_1)|\), ensuring that \(v\) does not lie inside \(S_2\). Thus, \(v_1, v_2\), and \(S_2\) satisfy the required conditions.
      \end{itemize}

    \item \textbf{Case 3:} \(k_2 - k_1 = 1\).  
    For any \(Q \subset Q'\) with \(v_Q \in \text{Del}_{k_1}(X)\) and \(v_{Q'} \in \text{Del}_{k_2}(X)\), we can write \(Q' = Q \cup \{v\}\). The condition \(v_Q \in \text{Del}_{k_1}(X)\) implies that there exists a ball \(D_1 \subset \mathbb{R}^3\) containing \(Q\), and all other points in \(X\) lie outside \(D_1\). Similarly, \(v_{Q'} \in \text{Del}_{k_2}(X)\) implies that there exists a ball \(D_2 \subset \mathbb{R}^3\) containing \(Q \cup \{v\}\), and all other points in \(X\) lie outside \(D_2\). The intersection \(D_1 \cap D_2\) is non-empty because \(Q \subset D_1 \cap D_2\). 

    We aim to prove the existence of a ball \(D\) such that \(D_1 \cap D_2 \subset D \subset D_1 \cup D_2\) and \(v\) lies on the boundary of \(D\). Let \(O_1\) and \(O_2\) be the centers of \(D_1\) and \(D_2\), respectively, and \(r_1\) and \(r_2\) their radii. We consider two subcases to complete the proof:

    - \textbf{Subcase 3.1:} \(D_1 \subset D_2.\)  
      First, consider the distance \(|l(O_2, v)|\): 
      
      - If \(|l(O_2, v)| \geq |l(O_2, O_1)| + r_1\), then the ball \(D\), centered at \(O_2\) with radius \(|l(O_2, v)|\), satisfies \(D_1 \subset D \subset D_2\), and \(v\) lies on the boundary of \(D\).
      
      - If \(|l(O_2, v)| < |l(O_2, O_1)| + r_1\), consider the farthest point \(P\) on \(D_1\) along the line segment \(l(O_2, O_1)\), such that \(|l(O_2, P)| = |l(O_2, O_1)| + r_1 > |l(O_2, v)|\). The perpendicular bisector of \(P\) and \(v\) intersects \(l(O_2, P)\) at a point \(O\). 
      
      We can prove that \(O\) lies on the line segment \(l(O_1, O_2)\) because \(v\) does not lie inside \(D_1\), implying \(|l(O_1, v)| \geq r_1 = |l(O_1, P)|\). Hence, the perpendicular bisector cannot intersect \(l(O_1, P)\) outside \(P\), and \(O\) must lie on \(l(O_1, O_2)\).

      Let \(D\) be the ball centered at \(O\) with radius \(|l(O, v)| = |l(O, P)|\). By construction, \(v\) lies on the boundary of \(D\), and \(D\) contains \(D_1\) because $O_1\in l(O,P)$ and $|l(O_1,P)|=r_1$. Furthermore, since \(O \in l(O_2, P)\) and \(|l(O_2, P)| \leq r_2\), we have \(D \subset D_2\). Thus, in this subcase, we can always find a ball \(D\) such that \(D_1 \cap D_2 = D_1 \subset D \subset D_2 = D_1 \cup D_2\), and \(v\) lies on the boundary of \(D\).

    - \textbf{Subcase 3.2:} \(D_1 \not\subset D_2.\)  
      We first prove the following lemma:  

      \textbf{Lemma.} For two intersecting balls \(D_1\) and \(D_2\) with centers \(O_1\) and \(O_2\), respectively, let \(C\) be the circle formed by the intersection of their boundaries. For any point \(O_3\) on the line segment \(l(O_1, O_2)\), the ball \(D\) centered at \(O_3\) with radius equal to the distance from \(O_3\) to any point \(P \in C\) satisfies \(D_1 \cap D_2 \subset D \subset D_1 \cup D_2\).  

      To prove this lemma, consider the plane \(\Pi\) containing \(P\) and the line segment \(l(O_1, O_2)\), and establish a 3D coordinate system on \(\Pi\):  
      - Let the projection of \(P\) onto \(l(O_1, O_2)\) be the origin \(O(0,0,0)\).  
      - Take the direction of \(O_1O_2\) as the \(x\)-axis, the direction of \(OP\) as the \(y\)-axis, and the direction orthogonal to \(\Pi\) as the \(z\)-axis.  
      - Let the coordinates of \(O_1\) and \(O_2\) be \((x_1,0,0)\) and \((x_2,0,0)\), respectively, and let \(P\) be at \((0, h, 0)\). Then, \(r_1^2 = x_1^2 + h^2\) and \(r_2^2 = x_2^2 + h^2\).  

      A point \((x, y, z)\) lies in \(D_1\) if and only if:  
      \[
      (x - x_1)^2 + y^2 + z^2 \leq r_1^2,
      \]
      which simplifies to:  
      \[
      x^2 - 2x_1x + y^2 + z^2 \leq h^2. \tag{1}
      \]  
      Similarly, a point lies in \(D_2\) if and only if:  
      \[
      x^2 - 2x_2x + y^2 + z^2 \leq h^2. \tag{2}
      \]  
      A point lies in \(D\) if and only if:  
      \[
      x^2 - 2(tx_1 + (1-t)x_2)x + y^2 + z^2 \leq h^2, \tag{3}
      \]
      where \(O_3(x_3, 0, 0)\) with \(x_3 = tx_1 + (1-t)x_2\) is the center of $D$.  

      By taking (1) \(\times t\) + (2) \(\times (1-t)\), we obtain (3). This implies \(D_1 \cap D_2 \subset D\). 

      Furthermore, the third inequality is equivalent to:  
\[
t(x^2 - 2x_1x + y^2 + z^2) + (1-t)(x^2 - 2x_2x + y^2 + z^2) \leq h^2.
\]  
Now, suppose that the third inequality holds while neither the first nor the second inequalities is satisfied. Then we have:  
\[
h^2 = th^2 + (1-t)h^2 < t(x^2 - 2x_1x + y^2 + z^2) + (1-t)(x^2 - 2x_2x + y^2 + z^2) \leq h^2,
\]  
which is a contradiction. This contradiction implies that if the third inequality holds, at least one of the first or second inequalities must also hold. This ensures that \(D \subset D_1 \cup D_2\). And we finish the proof of the lemma.

      Returning to Subcase 3.2, consider a point \(P \in C\) on the intersection of \(D_1\) and \(D_2\). Let \(\mathbf{H}\) be the perpendicular bisector of \(P\) and \(v\).  
      - Since \(|l(O_2, v)| \leq r_2 = |l(O_2, P)|\), \(\mathbf{H}\) intersects \(l(O_2, P)\).  
      - Since \(|l(O_1, v)| \geq r_1 = |l(O_1, P)|\), \(\mathbf{H}\) does not intersect \(l(O_1, P)\) internally.  

      By the geometric property that a line intersecting a triangle must intersect two sides and the fact that \(\mathbf{H}\) does not intersect \(l(O_1,P)\), \(\mathbf{H}\) intersects the triangle \(O_1O_2P\) at a point \(O\) on \(l(O_1, O_2)\). Let \(D\) be the ball centered at \(O\) with radius \(|l(O, P)| = |l(O, v)|\). By construction, \(v\) lies on the boundary of \(D\), and the lemma ensures \(D_1 \cap D_2 \subset D \subset D_1 \cup D_2\).  

    In conclusion, we have proven the existence of a ball \(D\) such that \(v\) lies on the boundary of \(D\), i.e., \(v \in \text{On}_X(S)\) where \(S = \partial D\). Furthermore, we have \(Q \subset D_1 \cap D_2 \subset D \subset D_1 \cup D_2\). This implies that \(Q = \text{In}_X(S)\), because the interiors of \(D_1\) and \(D_2\) do not contain any points of \(X\) other than those in \(Q \cup \{v\}\). By Theorem \ref{apxthm:basic}, it follows that \(v_Q \in C_{Q'}\).

\end{itemize}

\begin{theorem}
    Suppose \(X\) is in general position in \(\mathbb{R}^3\), and \(v_{Q_1}, v_{Q_2} \in \text{Del}_{k_1}(X),\, v_{Q'} \in \text{Del}_{k_2}(X)\). If \(v_{Q_1}, v_{Q_2} \in C_{Q'}\), then the following hold:
    \begin{enumerate}
        \item If \(3 \leq k_2 - k_1 \leq 4\), then \(Q_1 \subset Q'\) and \(N(Q_1, Q') \leq 5 - (k_2 - k_1)\).
        \item If \(Q_1 \not\subset Q'\), then \(N(Q_1, Q') \leq 3 - (k_2 - k_1)\).
        \item If \(Q_1 \cap Q' \subsetneq Q_2 \cap Q'\), then \(N(Q_1, Q') \leq N(Q_2, Q')\).
    \end{enumerate}
\end{theorem}

\textbf{Proof:} Let \(n_1 = |Q_1| - |Q' \cap Q_1|\) and \(n_2 = |Q'| - |Q' \cap Q_1|\). Then we have:

- \(n_1 \geq 0\), \(n_2 \geq 0\),

- \(n_2 - n_1 = k_2 - k_1 \leq 4\), because \(n_2 - n_1 = |Q'| - |Q_1| = k_2 - k_1\), and from Case 1 of Theorem \ref{apxthm:dim3}, we know \(k_2 - k_1 \leq 4\),

- \(n_2 + n_1 \leq 4\), because \(n_2 + n_1 = (|Q_1| + |Q'| - |Q' \cap Q_1|) - |Q' \cap Q_1| = |Q' \cup Q_1| - |Q' \cap Q_1|\). From Theorem \ref{apxthm:basic} and the fact that \(v_{Q_1} \in C(Q')\), we have \(|Q' \cup Q_1| - |Q' \cap Q_1| \leq |\text{In}_X(S) \cup \text{On}_X(S)| - |\text{In}_X(S)| = |\text{On}_X(S)| \leq 4\) for some sphere \(S \subset \mathbb{R}^3\).  

From these facts, we can enumerate all possible values for the pair \((n_2, n_1)\): \(\{(4,0), (3,0), (3,1), (2,0), (2,1), (1,0)\}\).

\begin{itemize}
    \item If \(3 \leq k_2 - k_1 \leq 4\), then \(Q_1 \subset Q'\) and \(N(Q_1, Q') \leq 5 - (k_2 - k_1)\).

    First, consider the case \(k_2 - k_1 = 4\). In this scenario, \(n_2 - n_1 = 4\). From the above analysis, we know \((n_2, n_1) = (4, 0)\), which implies \(|Q_1| - |Q' \cap Q_1| = 0\), i.e., \(Q_1 \subset Q'\). Therefore, \(v_{Q_1}\) and \(v_{Q'}\) can only lie in the depth-4 rhomboid \(\rho_X(S)\), where \(\text{In}_X(S) = Q_1\) and \(\text{On}_X(S) = Q' - Q_1\), meaning \(N(Q_1, Q') \leq 1\).

    Next, consider the case \(k_2 - k_1 = 3\). Similarly, we have \((n_2,n_1)=(3,0)\), which implies \(Q_1 \subset Q'\) and the fact that the minimal rhomboid \(\rho_S(X)\) containing \(v_{Q_1}\) and \(v_{Q'}\) could be either a depth-3 or depth-4 rhomboid with \((Q'-Q_1)\cup (Q_1-Q')\subset \text{On}_X(S)\). If \(\rho_S(X)\) is a depth-4 rhomboid, then \(N(Q_1, Q') \leq 1\). If \(\rho_S(X)\) is a depth-3 rhomboid, it can serve as a facet for at most two depth-4 rhomboids. Therefore, \(N(Q_1, Q') \leq 2\).

    \item If \(Q_1 \not\subset Q'\), then \(N(Q_1, Q') \leq 3 - (k_2 - k_1)\).

    From the above analysis, \(Q_1 \not\subset Q'\) implies \(k_2 - k_1 < 3\) and \(n_1 = |Q_1| - |Q' \cap Q_1| > 0\). First, consider the case \(k_2 - k_1 = 2\). Here, \(n_2 - n_1 = 2\), so \((n_2, n_1) = (3, 1)\). In this case, \(v_{Q_1}\) and \(v_{Q'}\) can only lie in the depth-4 rhomboid \(\rho_X(S)\), where \(\text{In}_X(S) = Q_1 \cap Q'\) and \(\text{On}_X(S) = (Q' - Q_1) \cup (Q_1 - Q')\). This means \(N(Q_1, Q') \leq 1\).

    Now, consider the case \(k_2 - k_1 = 1\). Similarly, \((n_2, n_1) = (2, 1)\). The minimal rhomboid \(\rho_S(X)\) containing \(v_{Q_1}\) and \(v_{Q'}\) can be either a depth-3 or depth-4 rhomboid. Thus, \(N(Q_1, Q') \leq 2\).

    \item If \(Q_1 \cap Q' \subsetneq Q_2 \cap Q'\), then \(N(Q_1, Q') \leq N(Q_2, Q')\).

    Since \(|Q_1| = |Q_2| = k_1\), the condition \(Q_1 \cap Q' \subsetneq Q_2 \cap Q'\) implies \(Q_1 \not\subset Q'\). From the above analysis, \(k_2 - k_1 < 3\), and \((n_2, n_1) = (3, 1)\) or \((2, 1)\). Let \(n_1' = |Q_2| - |Q' \cap Q_2|\) and \(n_2' = |Q'| - |Q' \cap Q_2|\). The condition \(Q_1 \cap Q' \subsetneq Q_2 \cap Q'\) implies \(n_1' = |Q_2| - |Q' \cap Q_2| < |Q_1| - |Q' \cap Q_1| = n_1 = 1\), which forces \(n_1' = 0\). Hence, \(Q_2 \subset Q'\).

    For any rhomboid \(\rho_X(S)\) containing \(v_{Q_1}\) and \(v_{Q'}\), we have \(\text{In}_X(S) \subset Q_1 \cap Q' \subsetneq Q_2 \cap Q'\), and \(Q_2 \cup Q' = Q' \subset Q_1 \cup Q' \subset \text{In}_X(S) \cup \text{On}_X(S)\). Therefore, \(\rho_X(S)\) also contains \(v_{Q_2}\). This implies \(N(Q_1, Q') \leq N(Q_2, Q')\).
\end{itemize}

\begin{theorem}
\label{thm:complexity}
Let \( K = \Delta k \cdot L + 1 \), where \( \Delta k \) is the step size and \( L \) is the number of pooling layers. Then the total time complexity of RTPool on a point cloud in $\mathbb{R}^d$ of size \( n \) is
\[
O\left(K^{\left\lceil \frac{d+3}{2} \right\rceil}n^{\left\lfloor \frac{d+1}{2} \right\rfloor}  + K^5 n^2 \right).
\]
The first term corresponds to the cost of constructing the rhomboid tiling up to order \( K \), and the second term accounts for the cumulative computation over all \( L \) pooling layers. Here, \( \lfloor \cdot \rfloor \) and \( \lceil \cdot \rceil \) denote the floor and ceiling functions, respectively.
\end{theorem}

\textbf{Proof:} The time complexity analysis of RTPool consists of two components: (1) constructing the rhomboid tiling structure from the input point cloud, and (2) performing graph pooling based on the rhomboid tiling.

\textbf{1. Rhomboid Tiling Construction.}  
According to Proposition 5 and Remark 8 in \cite{corbet2023computing}, the number of rhomboid cells in the tiling up to order \( K \) for a point cloud of size \( n \subset \mathbb{R}^d \) is bounded by
\[
O\left(K^{\left\lceil \frac{d+1}{2} \right\rceil}n^{\left\lfloor \frac{d+1}{2} \right\rfloor} \right).
\]
Furthermore, from \cite{corbet2021computing}, the time complexity of constructing a single order-\( k \) rhomboid cell is \( O(k) \). Therefore, the total time complexity for generating the rhomboid tiling up to order \( K \) is
\[
O\left(K\cdot K^{\left\lceil \frac{d+1}{2} \right\rceil} n^{\left\lfloor \frac{d+1}{2} \right\rfloor}\right) = O\left(K^{\left\lceil \frac{d+3}{2} \right\rceil} n^{\left\lfloor \frac{d+1}{2} \right\rfloor}  \right).
\]

\textbf{2. Graph Pooling via Rhomboid Tiling.}  
According to \cite{lee1982k}, the number of regions in the order-\( k \) Voronoi diagram in \( \mathbb{R}^3 \) is bounded by \( O(k^2(n - k)) \), and thus the number of vertices in the corresponding order-\( k \) Delaunay complex is also \( O(k^2(n - k)) \).

This implies that the clustering matrix constructed from the rhomboid tiling at order-\( k \) has size \( O(k^2(n - k)) \times O(k^2(n - k)) \), and the associated node feature matrix has size \( O(k^2(n - k)) \times O(1) \). The pooling process for each layer consists of two steps:
\begin{itemize}
    \item Matrix multiplication between the clustering matrix and the node feature matrix, with time complexity \( O(k^4 n^2) \).
    \item Applying a GIN layer to update the features, which has time complexity \( O(k^2(n - k)) \), assuming the graph remains sparse.
\end{itemize}
As the matrix multiplication dominates, the overall complexity per layer is \( O(k^4 n^2) \). Summing all orders up to \( K \), the total pooling cost becomes
\[
O(K \cdot K^4 n^2) = O(K^5 n^2).
\]

\textbf{Combining both parts}, the total time complexity of RTPool is
\[
O\left( K^{\left\lceil \frac{d+3}{2} \right\rceil}n^{\left\lfloor \frac{d+1}{2} \right\rfloor} + K^5 n^2 \right).
\]
This completes the proof.

\section{Hyperparameter Settings}
The hyperparameters of our RTPool model follow a default configuration, as shown in Table \ref{tab:default}. For different datasets, specific hyperparameters were adjusted to optimize performance, with the detailed settings presented in Table \ref{tab:dataset_hyperparameters}.

\begin{table}[htbp]
\centering
\caption{Default Hyperparameter Settings for RTPool}
\label{tab:default}
\begin{tabular}{|c|c|c|c|c|c|c|}
\hline
\texttt{batch size} & \texttt{\#epochs} & \texttt{LR} & \texttt{\#pooling layers} & $\Delta k$ & \texttt{final dropout} & \texttt{weight decay} \\ \hline
\texttt{16}         & \texttt{500}      & \texttt{0.001} & \texttt{2}   & \texttt{1}         & \texttt{0.5}           & \texttt{0.0001}        \\ \hline
\end{tabular}
\end{table}
The hyperparameters in Table \ref{tab:default} define the key settings for the RTPool model. The \texttt{batch size} specifies the number of samples processed simultaneously during training, while \texttt{\#epochs} represents the total number of complete passes through the training dataset. The \texttt{LR} (learning rate) controls the step size during the optimization process. The \texttt{\#pooling layers} denotes the number of pooling layers in the RTPool model, determining the hierarchical depth of the pooling process. $\Delta k$ is the step size in the
pooling process. The \texttt{final dropout} rate helps prevent overfitting by randomly zeroing out a fraction of neurons in the final layer. Lastly, the \texttt{weight decay} regularization term reduces the magnitude of model weights to improve generalization.

\begin{table}[htbp]
\centering
\caption{Hyperparameter Settings for Different Datasets}
\label{tab:dataset_hyperparameters}
\begin{tabular}{|l|c|c|c|c|c|}
\hline
\textbf{Dataset} & \texttt{\#pooling layers} & \texttt{LR}    & \texttt{final dropout} & $\Delta k$ & Others \\ \hline
COX2             & 1                            & 0.001          & 0.5                     & 2 & Default\\ \hline
BZR              & 1                            & 0.001          & 0.5                     & 1 &Default \\ \hline
PTC\_MR          & 1                            & 0.0002         & 0.3                     & 2 &Default \\ \hline
PTC\_MM          & 2                            & 0.0002         & 0.3                     & 1 &Default \\ \hline
PTC\_FR          & 2                            & 0.0002         & 0.3                     & 1 &Default \\ \hline
PTC\_FM          & 2                            & 0.0002         & 0.3                     & 1 &Default \\ \hline
MUTAG            & 1                            & 0.001          & 0.5                     & 1 &Default \\ \hline
\end{tabular}
\end{table}

In our experiments, we only tuned three hyperparameters: \texttt{\#pooling layers}, \texttt{LR}, and \texttt{final dropout}. For hyperparameter sensitivity analysis, we predefined a range of candidate values and performed grid search to identify the best-performing configuration on each dataset. Specifically, the optimal setting for each dataset was selected based on the validation performance, while other hyperparameters were kept at their default values.

\begin{table}[htbp]
    \centering
    \caption{Hyperparameter sensitivity analysis. Best performance is highlighted in \textbf{bold}.}
    
    % First row: Learning rate and Pooling layers
    \begin{minipage}[t]{0.32\textwidth}
        \centering
        \textbf{(a) Learning rate sensitivity} \\
        \begin{tabular}{@{}lcc@{}}
        \toprule
        \textbf{Dataset} & \texttt{LR} & \textbf{Accuracy} \\ 
        \midrule
        \multirow{4}{*}{COX2} & 0.0002 & 87.23±1.50 \\ 
                              & 0.0005 & 87.66±0.95 \\ 
                              & 0.001 & \textbf{92.76±1.90} \\ 
                              & 0.002 & 88.93±1.78 \\ 
        \midrule
        \multirow{4}{*}{MUTAG} & 0.0002 & 89.47±0.00 \\ 
                              & 0.0005 & 91.57±2.88 \\ 
                              & 0.001 & \textbf{94.74±3.33} \\ 
                              & 0.002 & 93.68±2.10 \\ 
        \midrule
        \multirow{4}{*}{PTC\_MR} & 0.0001 & 73.72±2.39 \\ 
                                & 0.0002 & \textbf{78.86±1.57} \\ 
                                & 0.0005 & 72.57±1.40 \\ 
                                & 0.001 & 70.29±1.40 \\ 
        \bottomrule
        \end{tabular}
        \label{table:lr}
    \end{minipage}
    \hfill
    \begin{minipage}[t]{0.32\textwidth}
        \centering
        \textbf{(b) \#pooling layers sensitivity} \\
        \begin{tabular}{@{}lc@{\hspace{0.5em}}c@{}}
        \toprule
        \textbf{Dataset} & \texttt{\#pooling} & \textbf{Accuracy} \\ 
                         & \texttt{layers} & \\ 
        \midrule
        \multirow{2}{*}{COX2} & 1 & \textbf{92.76±1.90} \\ 
                             & 2 & 88.50±1.17 \\ 
        \midrule
        \multirow{2}{*}{MUTAG} & 1 & \textbf{94.74±3.33} \\ 
                              & 2 & 90.52±2.35 \\ 
        \midrule
        \multirow{2}{*}{PTC\_MR} & 1 & \textbf{78.86±1.57} \\ 
                                & 2 & 72.57±2.56 \\ 
        \bottomrule
        \end{tabular}
        \label{table:pool_layers}
    \end{minipage}
    \hfill
    \begin{minipage}[t]{0.32\textwidth}
        \centering
        \textbf{(c) final dropout sensitivity} \\
        \begin{tabular}{@{}lc@{\hspace{0.5em}}c@{}}
        \toprule
        \textbf{Dataset} & \texttt{final} & \textbf{Accuracy} \\ 
                         & \texttt{dropout} & \\ 
        \midrule
        \multirow{3}{*}{COX2} & 0.4 & 88.51±1.91 \\ 
                             & 0.5 & \textbf{92.76±1.90} \\ 
                             & 0.6 & 88.93±1.78 \\ 
        \midrule
        \multirow{3}{*}{MUTAG} & 0.4 & 92.63±2.88 \\ 
                              & 0.5 & \textbf{94.74±3.33} \\ 
                              & 0.6 & 93.69±2.36\\ 
        \midrule
        \multirow{3}{*}{PTC\_MR} & 0.2 & 74.86±1.27 \\ 
                                & 0.3 & \textbf{78.86±1.57} \\ 
                                & 0.4 & 75.43±1.56\\ 
        \bottomrule
        \end{tabular}
        \label{table:dropout}
    \end{minipage}
    \label{OHSA}
\end{table}

\section{Hyperparameter Sensitivity Analysis}\label{HSA}

To evaluate the robustness of our method, we conducted a sensitivity analysis on four hyperparameters: the number of pooling layers (\texttt{\#pooling layers}), learning rate (\texttt{LR}), final dropout ratio (\texttt{final dropout}), and the step size $\Delta k$ in the
pooling process. While our main experiments only tuned the first three, we later performed additional evaluations of $\Delta k$ on three datasets (COX2, MUTAG, and PTC\_MR) to further investigate its impact.

The sensitivity analysis of the step size $\Delta k = k_2 - k_1$ is presented in the main text (see Table~\ref{table:delta_k}) along with a detailed discussion of its impact on model performance. Table~\ref{OHSA} reports the sensitivity results for the other three hyperparameters across the same datasets. 

In general, we find that our model RTPool is relatively stable across a range of values, but optimal performance does depend on careful tuning.

\section{RTPool for graph regresssion tasks}
We demonstrate that RTPool's applicability extends beyond graph classification to regression tasks, owing to its geometry-preserving pooling mechanism. By explicitly incorporating geometric information during pooling, RTPool generates graph-level representations that robustly capture essential structural features, making them universally applicable across diverse learning objectives.

To validate this claim, we evaluate RTPool on three commonly used molecular property regression benchmarks\cite{wu2018moleculenet} under identical experimental conditions as our main experiments. All models, including baselines, were tuned via grid search to ensure fair comparison. Table~\ref{tab:regression} presents the root mean square error (RMSE) comparisons with seven state-of-the-art graph pooling methods:

\begin{table*}[htbp]
\centering
\caption{Performance comparison (RMSE) on molecular property regression datasets under standardized experimental conditions. Lower values indicate better performance. Best results are highlighted in \textbf{bold}.}
\begin{tabular}{lccc}
\toprule
\textbf{Model} & \textbf{Esol} & \textbf{FreeSolv} & \textbf{Lipo} \\
No. graphs & 1128 & 642& 4200 \\
No. avg nodes & 26 & 18 & 49\\
\midrule
MinCutPool &2.1913±0.0374 &4.0111±0.0170 &1.3481±0.0224 \\
StructPool &2.1749±0.0411 & 4.0077±0.0150 &1.3422±0.1264 \\
DiffPool &3.7699±0.2035 & 5.2877±0.2049 &2.7431±0.0753 \\
HaarPool&2.1035±0.0340& 3.8892±0.0098& 1.3361±0.0627 \\
Wit-TopoPool &\textbf{1.8783±0.1628} &4.2159±0.0816 & 1.0916±0.5007 \\
Hop-Pool &2.4831±0.0760 &4.0030±0.0940 &1.3725±0.0738 \\
Mv-Pool &2.5691±0.0484 &4.0627±0.1048 &1.3746±0.0682 \\
\midrule
RTPool &2.0195±0.5318 &\textbf{3.6666±0.2112} &\textbf{1.0789±0.0496} \\
\bottomrule
\end{tabular}
\label{tab:regression}
\end{table*}

As shown in the table, RTPool achieves the best or near-best performance across all three datasets, demonstrating its effectiveness in graph regression tasks compared to other pooling baselines.

\section{RTPool on Social Network Datasets}

To extend the applicability of our RTPool model, we further explore its performance on non-geometric graphs—specifically, social network datasets such as IMDB-BINARY and IMDB-MULTI. These graphs lack explicit node features or spatial coordinates; thus, to adapt RTPool, a crucial step is embedding the nodes into a Euclidean space based on the graph structure alone.

\textbf{Spectral Embedding via Graph Laplacian.}
Given an unweighted, undirected graph $G = (V, E)$, we construct the normalized graph Laplacian $\mathcal{L} = I - D^{-1/2} A D^{-1/2}$, where $A$ is the adjacency matrix and $D$ is the diagonal degree matrix. We then compute the eigen-decomposition of $\mathcal{L}$ and select the eigenvectors corresponding to the three smallest non-zero eigenvalues (excluding the trivial eigenvector for eigenvalue zero). These three eigenvectors form a spectral embedding that maps each node to a point in $\mathbb{R}^3$, capturing global structural information through diffusion geometry. This embedding serves as a surrogate spatial coordinate input for RTPool, enabling it to perform pooling even in non-geometric graphs.

We applied this Laplacian-based spectral embedding approach to the IMDB-BINARY and IMDB-MULTI datasets, which contain only graph connectivity information. We compare our model with 20 baselines, including classical graph kernels (e.g., WL, HGK) and pooling-based GNNs (e.g., DiffPool, SAGPool, Wit-TopoPool), with most results cited from \cite{chen2023topological}. For Hop-Pool\cite{zhang2024multi} and Mv-Pool\cite{ma2024graphadt}, which were not reported in that benchmark, we implemented them based on their original papers and tuned hyperparameters via grid search to ensure fair comparison. The performance is summarized in Table~\ref{tab:social_results}.

\begin{table*}[htbp]
\centering
\caption{Comparison of classification accuracy (\%) on social network datasets.}
\begin{tabular}{lcc}
\toprule
\textbf{Model} & \textbf{IMDB-BINARY} & \textbf{IMDB-MULTI}\\
No. graphs & 1000 & 1500 \\
No. avg nodes & 19.77  & 13.00\\
\midrule
%CSM  & OOT & OOT  \\
HGK-SP   & 73.34±0.47 & 51.58±0.42  \\
HGK-WL  & 72.75±1.02 & 50.73±0.63 \\
WL & 71.15±0.47 & 50.25±0.72 \\
WL-OA  & 74.01±0.66 & 49.95±0.46  \\
DGCNN  & 70.00±0.90 & 47.80±0.90\\
GCN & 66.53±2.33 & 48.93±0.88  \\
GIN  & 75.10±5.10 & 52.30±2.80  \\
Top-$K$ & 73.17±4.84 & 48.80±3.19  \\
MinCutPool & 70.77±4.89 & 49.00±2.83  \\
DiffPool  & 68.60±3.10 & 45.70±3.40  \\
EigenGCN  & 70.40±1.30 & 47.20±3.00\\
SAGPool & 74.87±4.09 & 49.33±4.90  \\
HaarPool & 73.29±3.40 & 49.98±5.70 \\
PersLay  & 71.20±0.70 & 48.80±0.60  \\
FC-V  & 73.84±0.36 & 46.80±0.37  \\
MPR  & 73.80±4.50 & 50.90±2.50\\
SIN  & 75.60±3.20 & 52.50±3.00 \\
Wit-TopoPool  & \textbf{78.40±1.50} & \textbf{53.33±2.47} \\
Hop-Pool &68.04±2.04  & 49.33±5.09\\
Mv-Pool &69.75±3.61  & 51.67±0.74 \\
\midrule
RTPool &73.06±3.84 &\textbf{53.33±1.26} \\
\bottomrule
\end{tabular}
\label{tab:social_results}
\end{table*}

As shown in Table~\ref{tab:social_results}, RTPool achieves competitive results across both datasets. On IMDB-MULTI, RTPool matches the best-performing baseline (Wit-TopoPool) with an accuracy of 53.33\%. Notably, this is achieved despite the lack of native geometric information, highlighting the effectiveness of using spectral embeddings to generalize RTPool to abstract graph domains like social networks.

\section{Empirical Model Efficiency} \label{sec:efficiency}

To further evaluate the efficiency of RTPool, we compare it with several representative clustering-based pooling methods under controlled training conditions. Specifically, all models are trained for only 100 epochs per run (rather than the default 500), and each experiment is repeated five times to report average performance and runtime. This setup helps assess how quickly each model converges to high-quality results.

\paragraph{Early-Stage Performance.}
Table~\ref{table:result} shows the classification performance of different pooling methods after only 100 training epochs. Despite the reduced training time, RTPool already achieves outstanding accuracy on most datasets, often surpassing or closely matching state-of-the-art models that typically require longer training. This demonstrates the rapid convergence and high early-stage expressiveness of our model.

\begin{table*}[htbp]
\centering
\caption{Accuracy (mean ± std) after only 100 training epochs. Each model is trained 5 times. RTPool shows strong early-stage performance compared to other methods.}
\begin{tabular}{lccccccc}
\toprule
\textbf{Model} & \textbf{BZR} & \textbf{COX2} & \textbf{MUTAG} & \textbf{PTC\_MR} & \textbf{PTC\_MM} & \textbf{PTC\_FM} & \textbf{PTC\_FR} \\
\midrule
MinCutPool & 76.47±2.32 & 79.86±2.47 & 69.47±2.11 & 66.86±2.91 & 72.94±1.18 & 56.74±5.63 & 62.86±1.81 \\
StructPool & 75.63±1.04 & 78.72±1.79 & 70.53±2.58 & 65.74±2.68 & 72.90±1.18 & 57.71±7.32 & 62.29±2.14 \\
DiffPool & 78.54±0.98 & 77.93±3.18 & 73.68±3.33 & 69.14±2.80 & 67.06±2.20 & 68.57±3.61 & \textbf{68.57±1.81} \\
HaarPool & 78.05±0.00 & 80.64±4.58 & 68.42±0.38 & 62.29±4.20 & 64.71±3.72 & 61.14±6.66 & 65.67±2.25 \\
Wit-TopoPool & 80.98±2.39 & 80.43±1.59 & 85.32±2.58 & 71.84±1.14 & 72.82±4.71 & 68.56±4.84 & \textbf{68.57±4.04} \\
Hop-Pool & 78.05±1.39 & 80.00±1.70 & \textbf{87.56±4.41} & 61.71±2.91 & 70.59±0.78 & 58.29±1.40 & 63.43±1.14 \\
Mv-Pool & 75.60±1.98 & 79.68±1.27 & 73.68±10.53 & 64.57±4.64 & 68.24±1.18 & 57.14±0.00 & 65.71±1.77 \\
\midrule
RTPool & \textbf{84.39±1.19} & \textbf{85.96±1.04} & 83.16±5.16 & \textbf{72.86±3.65} & \textbf{72.94±3.43} & \textbf{68.86±8.59} & 67.71±7.75 \\
\bottomrule
\end{tabular}
\label{table:result}
\end{table*}

\paragraph{Runtime Comparison.}
Table~\ref{table:time} presents the total training time (in seconds) under the same 100-epoch setup. For RTPool, we additionally break down the cost into the one-time tiling structure construction and model training time. While some methods incur large computation costs (e.g., Mv-Pool, Wit-TopoPool), RTPool maintains a reasonable runtime and even outperforms most baselines in both speed and accuracy.

\begin{table*}[htbp]
\centering
\caption{Runtime (in seconds) over 5 trials (each with 100 epochs). For RTPool, both tiling construction time and training time are reported.}
\begin{tabular}{lccccccc}
\toprule
\textbf{Model} & \textbf{BZR} & \textbf{COX2} & \textbf{MUTAG} & \textbf{PTC\_MR} & \textbf{PTC\_MM} & \textbf{PTC\_FM} & \textbf{PTC\_FR} \\
\midrule
MinCutPool & 2297.74 & 2554.96 & 1999.89 & 2016.22 & 2482.84 & 2470.79 & 2355.13 \\
StructPool & 1602.96 & 1647.23 & 1387.76 & 1412.16 & 1663.12 & 1623.40 & 1578.65 \\
DiffPool & 2507.53 & 2813.80 & 2144.75 & 2235.34 & 2637.06 & 2621.80 & 2529.79 \\
HaarPool & 3238.15 & 2666.93 & 1787.79 & 1412.01 & 1407.39 & 2075.04 & 2022.91 \\
Wit-TopoPool & 4512.64 & 4904.08 & 7475.70 & 7390.13 & 7332.05 & 7370.51 & 7357.56 \\
Hop-Pool & 1483.80 & 1420.63 & 1190.00 & 1171.09 & 1450.05 & 1421.54 & 1355.92 \\
Mv-Pool & 11244.74 & 9137.48 & 5828.78 & 8935.74 & 8327.48 & 8326.58 & 8907.54 \\
\midrule
RTPool (constructor) & 1616.06 & 2356.41 & 189.59 & 341.39 & 321.39 & 331.57 & 360.00 \\
RTPool (training) & 1117.62 & 1053.07 & 1041.59 & 673.77 & 1268.01 & 1708.35 & 306.10 \\
\bottomrule
\end{tabular}
\label{table:time}
\end{table*}

These results highlight RTPool’s fast convergence and favorable runtime-accuracy trade-off, making it a practical and scalable choice for large-scale or resource-constrained applications.

\section{Illustrative Examples}\label{exp:rhomboidtiling}

To provide a more intuitive understanding of RTpool and the rhomboid tiling mechanism, we include two illustrative examples in this appendix.

\begin{figure*}[htbp]
\centering
\includegraphics[width=0.7\linewidth]{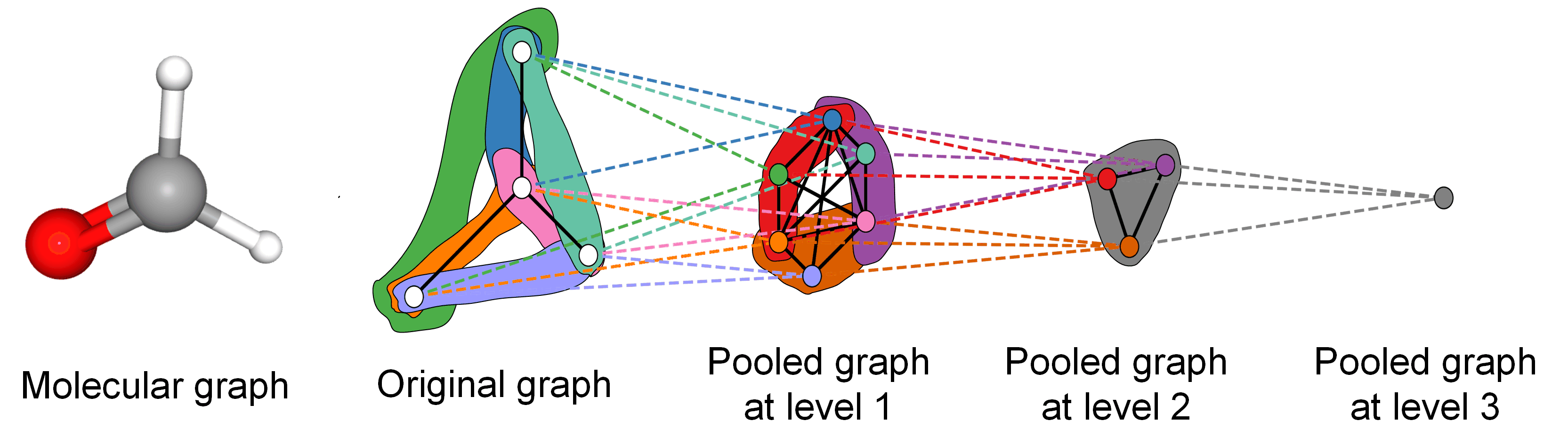}
\caption{Visualization of the hierarchical clustering process performed by RTpool on the molecular graph of Formaldehyde. The original molecular structure is shown on the left, followed by three successive clustering layers. Each cluster center (colored circle) is positioned based on the geometric realization of rhomboid tiling, and is connected to its members by dashed lines.}
\label{fig:RTpool_visual}
\end{figure*}

Figure~\ref{fig:RTpool_visual} demonstrates how RTpool performs hierarchical clustering on the molecular graph of Formaldehyde. The graph is progressively pooled through multiple layers, with nodes being aggregated into clusters at each level. The position of each cluster center is determined by the rhomboid tiling structure, providing geometric guidance for the pooling process. This example reveals RTpool’s strong locality-awareness, as it consistently merges spatially adjacent nodes. In this example, we observe that RTpool tends to group geometrically close nodes into the same cluster, a phenomenon particularly evident from pooling layer 1 to layer 2.

\begin{figure*}[htbp]
\centering
\includegraphics[width=0.95\linewidth]{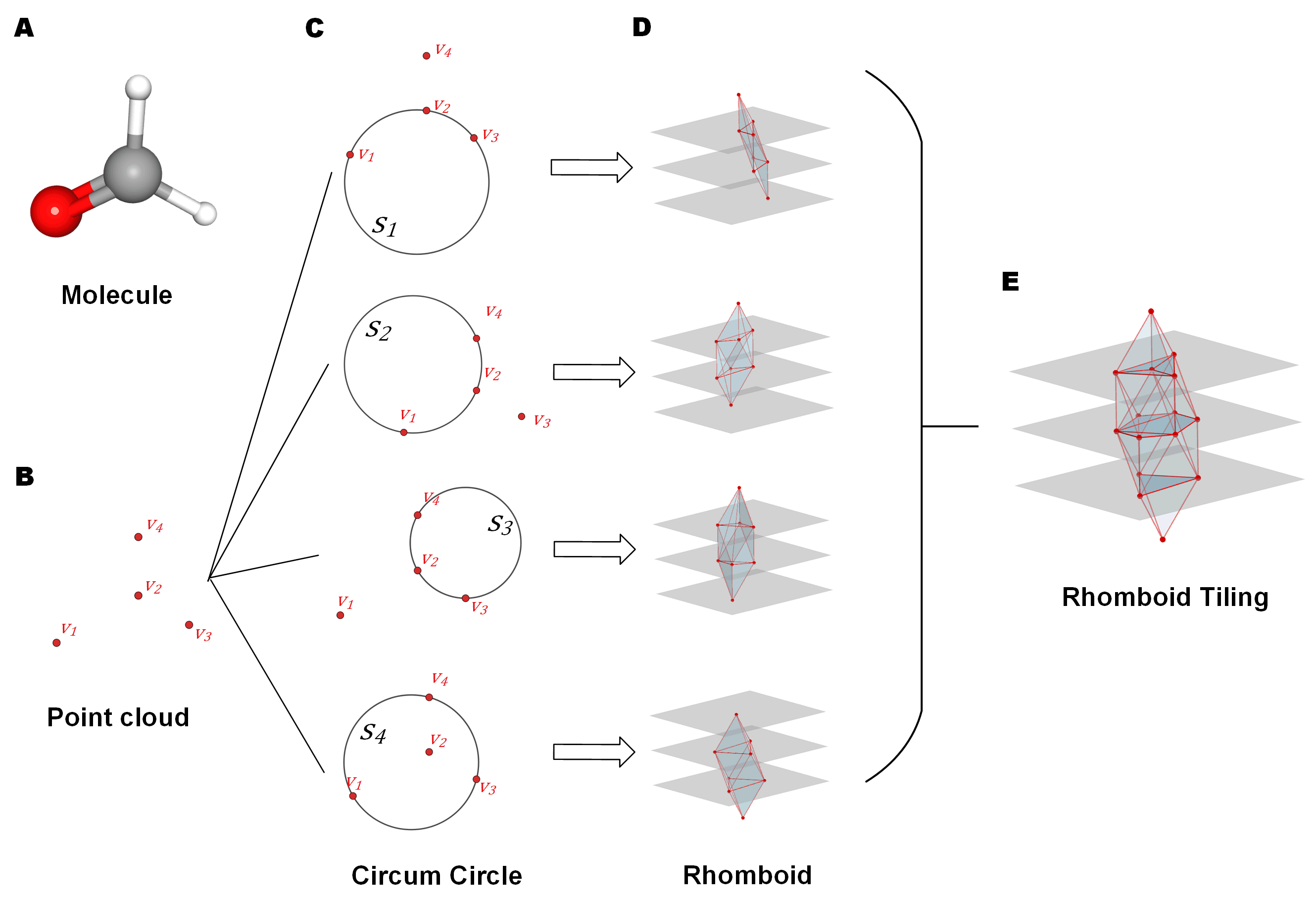}
\caption{An illustrative example of generating a rhomboid tiling from a point cloud. \textbf{A}: Molecular graph of Formaldehyde. \textbf{B}: Point cloud obtained by embedding the molecular graph into $\mathbb{R}^2$. \textbf{C}: Minimal circumcircles corresponding to each local point cluster in the point cloud. \textbf{D}: Rhomboids associated with the circumcircles. \textbf{E}: Rhomboid tiling structure of Formaldehyde formed by the union of all rhomboids.}
\label{fig:RT_construction}
\end{figure*}

Figure~\ref{fig:RT_construction} illustrates the process of constructing a rhomboid tiling from a 2D point cloud $X$ derived from the molecular graph of Formaldehyde. We begin by identifying minimal circumcircles that pass through exactly three points. These circles are used to define the rhomboids that make up the tiling.

Taking one such minimal circumcircle $S_1$ as an example, suppose $\text{In}_X(S_1) = \emptyset$ and $\text{On}_X(S_1) = {v_1, v_2, v_3}$. With $(a_i, b_i)$ denoting the coordinates of $v_i$, the associated rhomboid is formed from the convex hull of a set of lifted points:

\[
\rho_X(S_1) := \text{conv}\{y_Q \mid \text{In}_X(S_1) \subset Q \subset \text{In}_X(S_1) \cup \text{On}_X(S_1)\},
\]
\[
y_Q := \left(\sum_{x \in Q} x,\ -k\right)
\]

In this case, we generate 8 such points corresponding to all subsets $Q$ of $\{v_1, v_2, v_3\}$, at levels ranging from 0 to 3:
\begin{itemize}
    \item $y_\emptyset = (0, 0, 0)$ (origin, at level 0),
    \item $y_{\{v_1\}} = (a_1, b_1, -1)$, $y_{\{v_2\}} = (a_2, b_2, -1)$, $y_{\{v_3\}} = (a_3, b_3, -1)$ (three points at level 1),
    \item $y_{\{v_1,v_2\}}=(a_1+a_2,b_1+b_2,-2)$, $y_{\{v_1,v_3\}}=(a_1+a_3,b_1+b_3,-2)$, $y_{\{v_2,v_3\}}=(a_2+a_3,b_2+b_3,-2)$ (three points at level 2),
    \item $y_{\{v_1,v_2,v_3\}} = (a_1+a_2+a_3, b_1+b_2+b_3, -3)$ (one point at level 3).
\end{itemize}
The convex hull of these 8 lifted points defines a rhomboid in $\mathbb{R}^3$. As shown in the figure, four such rhomboids are constructed from four distinct minimal circumcircles. Their union forms the complete rhomboid tiling structure for the Formaldehyde point cloud.

\end{document}